\documentclass[sigconf]{acmart}
\AtBeginDocument{%
  }

\copyrightyear{2025}
\acmYear{2025}
\setcopyright{acmlicensed}
\acmConference[MM '25] {Proceedings of the 33rd ACM International Conference on Multimedia}{October 27--31, 2025}{Dublin, Ireland.}
\acmBooktitle{Proceedings of the 33rd ACM International Conference on Multimedia (MM '25), October 27--31, 2025, Dublin, Ireland}
\acmDOI{10.1145/3746027.3755203}
\acmISBN{979-8-4007-2035-2/2025/10}

\acmSubmissionID{2584}

\usepackage{xcolor}
\usepackage{makecell}
\usepackage{tcolorbox}
\usepackage[linesnumbered,vlined,boxed,commentsnumbered,ruled,norelsize]{algorithm2e}
\usepackage{multirow}
\usepackage{booktabs}
\usepackage{soul}
\usepackage{threeparttable}
\usepackage{colortbl,xcolor}
\usepackage{makecell}
\usepackage{balance}
\usepackage{titlesec}
\usepackage{tcolorbox}
\usepackage{subfigure}
\usepackage{amsmath}
\usepackage{balance}
\usepackage{enumitem}

\begin{document}

\newcommand\vldbdoi{10.1145/3746027.3755203}
\newcommand\vldbpages{XXX-XXX}
\newcommand\vldbvolume{14}
\newcommand\vldbissue{1}
\newcommand\vldbyear{2020}
\newcommand\vldbauthors{\authors}
\newcommand\vldbtitle{\shorttitle} 
\newcommand\vldbavailabilityurl{URL_TO_YOUR_ARTIFACTS}
\newcommand\vldbpagestyle{empty}

\definecolor{LightPurple}{RGB}{230,230,247}   
\definecolor{mygreen}{RGB}{0,176,80}
\definecolor{myblue}{rgb}{0.267, 0.447, 0.769}
\definecolor{myred}{RGB}{192,0,0}
\definecolor{TableGreen}{RGB}{0,176,80}
\definecolor{TableRed}{RGB}{161,0,0}
\definecolor{LightBlue}{RGB}{79,97,209}
\definecolor{TableDarkRed}{RGB}{192,0,0}
\definecolor{TableDarkGreen}{RGB}{114,177,39}
\definecolor{TableGray}{RGB}{236,236,236}

\newcommand{\wuhao}{\fontsize{8pt}{8pt}\selectfont}
\newcommand{\xiaowuhao}{\fontsize{7.7pt}{7.7pt}\selectfont}
\newcommand{\sizetwo}{\fontsize{18.5pt}{18.5pt}\selectfont}
\newcommand{\sizethree}{\fontsize{12.5pt}{12.5pt}\selectfont}
\newcommand{\sizesmallthree}{\fontsize{9.6pt}{9.6pt}\selectfont}
\newcommand{\sizefour}{\fontsize{9pt}{9pt}\selectfont}
\newcommand{\sizemidfour}{\fontsize{8.8pt}{8.8pt}\selectfont}
\newcommand{\sizesmallfour}{\fontsize{8.5pt}{8.5pt}\selectfont}
\newcommand{\sizefive}{\fontsize{8pt}{8pt}\selectfont}
\newcommand{\sizemidfive}{\fontsize{7.95pt}{7.95pt}\selectfont}
\newcommand{\sizesmallfive}{\fontsize{7.7pt}{7.7pt}\selectfont}
\newcommand{\sizetable}{\fontsize{7.2pt}{7.2pt}\selectfont}
\newcommand{\sizesix}{\fontsize{7pt}{7pt}\selectfont}
\newcommand{\sizesmallsix}{\fontsize{6.7pt}{6.7pt}\selectfont}
\newcommand{\sizeseven}{\fontsize{6.3pt}{6.3pt}\selectfont}
\newcommand{\sizesmallseven}{\fontsize{6.0pt}{6.0pt}\selectfont}
\newcommand{\erhao}{\fontsize{22pt}{22pt}\selectfont}

\title{\textsc{CapRecover}: A Cross-Modality Feature Inversion Attack Framework on Vision Language Models}



\author{Kedong Xiu}
\orcid{0009-0007-6409-9168}
\authornote{Also with Zhejiang University. Xiu finished this work when he was a remote intern at New York University.}
\affiliation{%
  \institution{New York University}
  \city{New York}
  \state{NY}
  \country{USA}
}
\email{kedongxiu@zju.edu.cn}

\author{Sai Qian Zhang}
\orcid{0000-0002-4815-9235}
\authornote{Corresponding author.}
\affiliation{%
  \institution{New York University}
  \city{New York}
  \state{NY}
  \country{USA}}
\email{sai.zhang@nyu.edu}






\renewcommand{\shortauthors}{Kedong Xiu and Sai Qian Zhang}

\begin{abstract}
 

As Vision-Language Models (VLMs) become increasingly integrated into user-facing applications, they are often deployed in split DNN configurations, where the visual encoder (e.g., ResNet or ViT) runs on user-side devices and only intermediate features are transmitted to the cloud for downstream processing. While this setup reduces communication overhead, the intermediate data features containing sensitive information can also expose users to privacy risks. Prior work has attempted to reconstruct images from these features to infer semantics, but such approaches often produce blurry images that obscure semantic details. In contrast, the potential to directly recover high-level semantic content — such as image labels or captions — via a cross-modality inversion attack remains largely unexplored. To address this gap, we propose \textsc{CapRecover}, a general cross-modality feature inversion framework that directly decodes semantic information from intermediate features without requiring image reconstruction. Additionally, \textsc{CapRecover} can be used to reverse engineer traditional neural networks for computer vision tasks, such as ViT, ResNet, and others.



We evaluate \textsc{CapRecover} across multiple widely used datasets and victim models. Our results demonstrate that \textsc{CapRecover} can accurately recover both image labels and captions without reconstructing a single pixel. Specifically, it achieves up to 92.71\% Top-1 accuracy on the CIFAR-10 dataset for label recovery, and generates fluent and relevant captions from ResNet50's intermediate features on COCO2017 dataset, with ROUGE-L scores up to 0.52. Furthermore, an in-depth analysis of ResNet-based models reveals that deeper convolutional layers encode significantly more semantic information, whereas shallow layers contribute minimally to semantic leakage. Furthermore, we propose a straightforward and effective protection approach that adds random noise to the intermediate image features at each middle layer and subsequently removes the noise in the following layer. Our experiments indicate that this approach effectively prevents information leakage without additional training costs. Our code is available \href{https://jus1mple.github.io/Image2CaptionAttack}{\textcolor{blue}{\textit{here}}}.


\end{abstract}





\maketitle


\section{Introduction}
\label{sec:introduction}

\par\noindent The rapid advancement of Vision-Language Models (VLMs) has fundamentally reshaped the landscape of multimodal AI, positioning these models as the cornerstone of modern user-facing assistants. Unlike traditional Large Language Models (LLMs), VLMs seamlessly integrate image understanding with natural language processing, enabling comprehensive interpretations of real-world data. By harnessing vast amounts of textual and visual information, VLMs have achieved impressive results in tasks such as image captioning (e.g., GPT-4o \cite{openai_gpt4o}), text-to-image generation (e.g., Stable Diffusion \cite{stable_diffusion}), and optical character recognition. The success of architectures like CLIP \cite{clip} and BLIP2 \cite{blip2} underscores their potential to drive significant innovations in both research and practical applications.

\begin{figure*}[thbp]
    \centering
    \includegraphics[width = 0.88\textwidth]{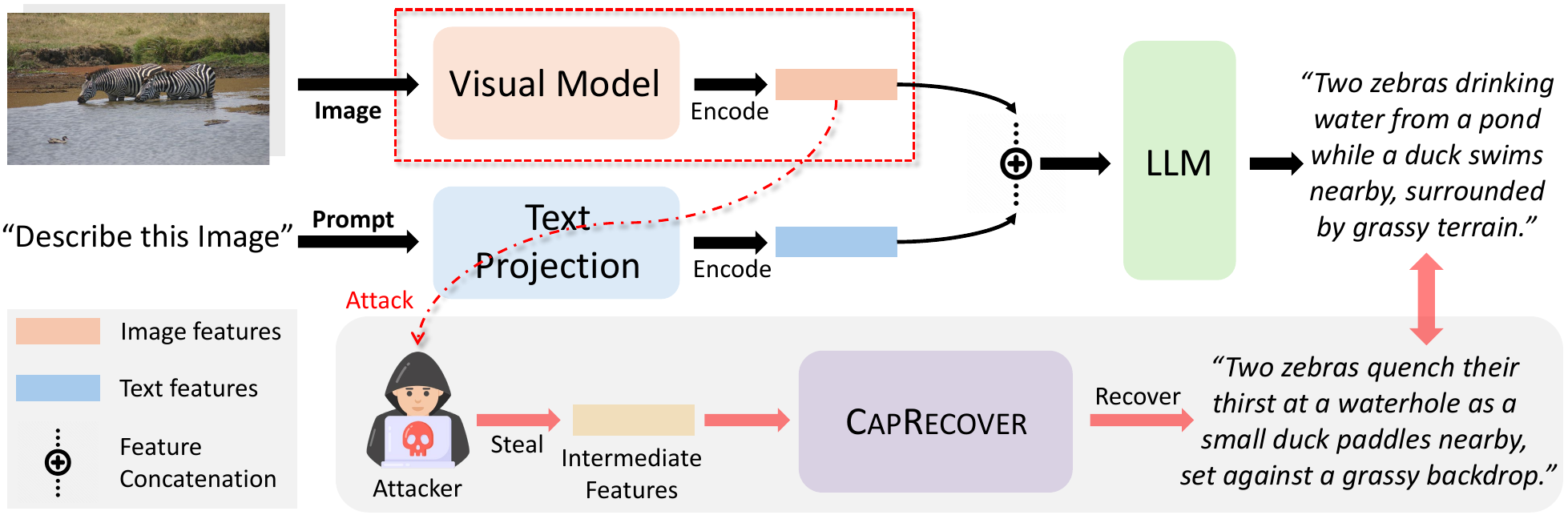}
    \vspace{-0.9em}
    \caption{Illustration of the cross-modality feature inversion attack scenario. In the depicted attack scenario, the adversary steals the intermediate image features from the visual model. Leveraging these stolen features, the adversary employs \textsc{CapRecover} to reconstruct the image caption/label, potentially revealing sensitive or private information.}
    \label{fig:illustration}
    \vspace{-0.5em}
\end{figure*}

Despite these advances, VLMs are not without vulnerabilities. Recent research has predominantly focused on security threats such as prompt jailbreaking—where attackers manipulate models to produce harmful outputs—and prompt-stealing attacks that extract sensitive user prompts from generated images \cite{gong2023figstep,luo2024jailbreakv28k,shayegani2023plug,2024_usenix_prompt_stealing_attack}. However, one critical dimension remains underexplored: the leakage of sensitive information through intermediate feature representations. As illustrated in Figure \ref{fig:illustration}, an adversary who gains access to intermediate image features from the victim model (for instance, from a local device) could reconstruct the original image caption, potentially exposing private user data. This issue is especially relevant in the split DNN computing paradigm~\cite{mudvari2024splitllm,zhang2024edgeshard,he2024large,lu2024merge}, where a large model is divided into multiple blocks tailored to the computational capabilities of edge devices. In this setup, user data is initially processed on the edge device using the first layers of the model, and intermediate results are transmitted to a remote server for processing by later layers. This data transfer poses a risk, as it can be intercepted and exploited by attackers to reconstruct user inputs. Consequently, understanding and mitigating the leakage of intermediate image features is imperative for protecting user data and maintaining the integrity of VLM-driven services.

Prior work \cite{xu2024stealthy, zhu2025passiveinfer, he2019model} has explored reconstructing images from intermediate features to further infer their semantic content. However, these methods are indirect and often suffer from preserving fine-grained semantics with poor visual fidelity (e.g., blurriness, missing textures), which could consequently limit their performance. Moreover, some attackers may primarily focus on the semantic meaning in the target image, e.g., what is happening and who is involved. This raises a critical yet underexplored question: \emph{Is it possible—and potentially more effective—for an attacker to directly recover high-level semantic information, such as image labels or captions, from intermediate features, without reconstructing the image at all?} This new form of \textit{feature inversion attack} shifts the adversary's focus from pixel-level recovery to semantic reconstruction, and more directly threatens user privacy in practical scenarios.


\subsection{Our work and contributions}
\label{subsec:intro_our_work_contri}


In this paper, we take a fundamentally different approach: instead of reconstructing the image, we \emph{directly recover/reconstruct the image's semantic content} from the leaked intermediate image features. We introduce \textsc{CapRecover}, a generic cross-modality feature inversion framework that exposes a critical vulnerability in VLMs: the capacity to reconstruct textual descriptions from intermediate image features. \textsc{CapRecover} bridges intermediate visual features with a pre-trained language model, bypassing image reconstruction entirely. By learning a lightweight projection layer between the vision and language domains, \textsc{CapRecover} enables accurate and fluent semantic recovery from commonly used encoders such as ResNet and ViT.

To understand the privacy implications of this attack, we consider a threat model where the attacker passively observes the intermediate visual features sent from a user's device to the cloud in a split-VLM pipeline. The attacker has no access to the original image or the language module of the VLM, and aims to infer semantic content directly from the encoder output. While a full description is provided in Sec. \ref{sec:threat_model}, we note here that this threat model aligns with realistic deployment scenarios in edge-cloud systems and highlights a previously underestimated attack surface.

%



We evaluate \textsc{CapRecover} on multiple datasets and VLM architectures across two key tasks: image classification and image captioning. Our experiments show that \textsc{CapRecover} can recover labels and captions with high fidelity—even without reconstructing a single pixel. We further analyze how semantic leakage varies across encoder depths and propose a simple, training-free defense mechanism that reduces leakage via reversible noise injection.

We summarize our contributions as follows:

\begin{itemize}
    
    

    \item \textbf{A general adversarial framework.} We propose \textsc{CapRecover}, the first generic cross-modality feature inversion framework that directly reconstructs semantic information from the leaked intermediate image features, without requiring pixel-level image reconstruction. By leveraging a feature-to-text alignment mechanism, \textsc{CapRecover} effectively recovers image labels and captions even in the absence of explicit textual outputs.

    \item \textbf{Extensive evaluation and analysis.} We evaluate \textsc{CapRecover} on both image classification and image captioning tasks using widely adopted datasets and victim models. \textsc{CapRecover} achieves up to 92.71\% Top-1 accuracy on CIFAR-10 for label recovery and a ROUGE-L score of 0.52 on COCO2017 for caption reconstruction.

    \item \textbf{An effective protection approach.} We propose a straightforward yet effective protection approach: Add random noise to the output of each layer in the victim model and remove this noise in the subsequent layer. Our approach only needs a small noise cost without any additional training cost, which can effectively mitigate the risks of sensitive information leakage from the intermediate image features.

\end{itemize}




\section{Threat Model}
\label{sec:threat_model}

We consider a cross-modality feature inversion attack scenario where an adversary aims to reconstruct/recover the semantic description/label corresponding to a given image by exploiting the \textit{intermediate image features} $\mathbf{F}$ produced by the victim visual encoder $\mathcal{V}_{image}$. We assume a reasonable deployment situation consistent with practical deployment \cite{he2024large, lu2024merge} where VLMs are deployed in user-facing applications or on edge devices, which commonly keep raw images and final captions locally private, yet may expose intermediate features (e.g., when features are transmitted to a cloud service or temporarily stored in device memory).

\begin{figure}[t]
    \centering
    \includegraphics[width = 0.93\linewidth]{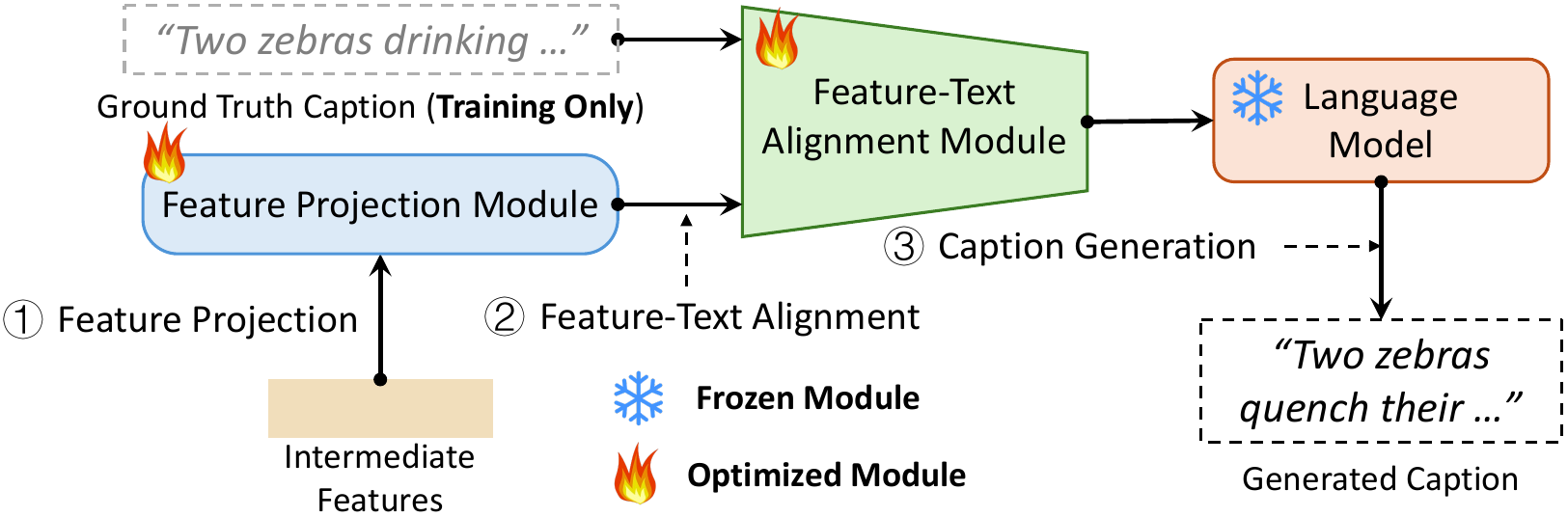}
    \vspace{-0.8em}
    \caption{Overview of \textsc{CapRecover}. \textsc{CapRecover} mainly consists of: (1) Feature projection module, (2) Feature-text alignment module, and (3) Caption generation module. We freeze the language model and optimize other modules.}
    \label{fig:model_overview}
    \vspace{-0.8em}
\end{figure}

\begin{table}[t]
    \centering
    \caption{Datasets Used in this paper.}
    \label{tab:dsts_used}
    \vspace{-1.0em}
    \begin{threeparttable}
    
    \sizefive
    \setlength{\tabcolsep}{4pt} 
    \renewcommand{\arraystretch}{1.1} 
    \begin{tabular}{crrrr}
    \toprule
    \textbf{Dataset}{$^\dagger$} & \makecell[c]{\textbf{Training size}} & \makecell[c]{\textbf{Sample size}} & \makecell[c]{\textbf{Test size}} & \makecell[c]{\textbf{Sample size}} \\ 
    \midrule
    COCO2017 & 118,287 & 30,000 & 5,000 & 5,000 \\
    Flickr8K & 6,000 & 6,000 & 1,000 & 1,000 \\
    ImageNet-1K & 1,281,167 & 12,000 & 50,000 & 900 \\
    \hline
    CIFAR-10 & 50,000 & 50,000 & 10,000 & 10,000 \\
    TinyImageNet & 100,000 & 100,000 & 10,000 & 10,000 \\
    \bottomrule
    \end{tabular}
    \begin{tablenotes}
        \item \hspace{-1.2em}$^{\dagger}$ {We use CIFAR-10 and TinyImageNet datasets for image label recovery and COCO2017, Flickr8K and ImageNet-1K datasets for image caption reconstruction, respectively.}
    \end{tablenotes}
    \end{threeparttable}
    \vspace{-0.9em}
\end{table}

\subsection{Adversary's Capabilities and Access.} 
\label{subsec:adv_cap}


We assume the adversary can intercept or obtain the victim model’s intermediate visual representations $\mathbf{F}$ but \emph{can not} directly access to the original input image $I$ or its corresponding semantic description (e.g., the ground truth caption $T_{cap}$ and image label $y_{cls}$). This can occur in practical scenarios where:
\begin{itemize}[leftmargin=1.0em]
    \item The adversary intercepts intermediate features transmitted from an edge device to a cloud server responsible for caption or label generation;
    \item A malicious insider or malware on the user’s device extracts intermediate features from memory.
\end{itemize}
We assume that the attacker knows the architecture of the victim’s visual encoder (e.g., ResNet50, ViT), including the position of intermediate layers used for downstream tasks. The attacker may also leverage auxiliary resources such as publicly available pretrained models to assist in decoding the extracted features. This setup aligns with a standard \textit{white-box} or \textit{gray-box} threat model.


\subsection{Adversary's Objective.} 
\label{subsec:adv_obj}


The adversary's goal is to exploit the intermediate image features $\mathbf{F}$ produced by the victim visual encoder ($\mathbf{F} = \mathcal{V}_{Image}(I)$) to reconstruct high-level semantic information. In this paper, we mainly consider two forms of semantic targets:
\begin{itemize}[leftmargin=1.0em]
    \item \textbf{Caption Reconstruction:} The attacker trains a cross-modality inversion attack model $\mathcal{A}_\theta$ to generate a textual caption $T'_{\text{cap}} = \mathcal{A}_\theta(\mathbf{F})$ that approximates the ground-truth caption $T_{\text{cap}}$;
    \item \textbf{Label Recovery:} The attacker trains $\mathcal{A}_\theta$ as a classifier to predict the image label $y'_{\text{cls}} = \mathcal{A}_\theta(\mathbf{F})$ matching the true label $y_{\text{cls}}$.
\end{itemize}

For caption reconstruction, the attacker minimizes a semantic loss between the generated and reference captions:
\begin{equation}
    \arg\min\limits_{\theta} \; \mathcal{L}(T'_{\text{cap}}, T_{\text{cap}}), \label{eq:caption_obj}
\end{equation}
where $\mathcal{L}(\cdot,\cdot)$ is a semantic loss function (e.g., based on token-level or embedding-level similarity; see Sec.~\ref{sec:method} for details).

For label recovery, the objective reduces to a standard classification loss:
\begin{equation}
    \arg\min\limits_{\theta} \; \mathcal{L}_{\text{cls}}(y'_{\text{cls}}, y_{\text{cls}}). \label{eq:label_obj}
\end{equation}

A successful attack implies that even without accessing raw pixels, the intermediate image features alone are sufficient to compromise user privacy by revealing semantic-level information.

\section{Methodology of \textsc{CapRecover}}
\label{sec:method}

\par\noindent In this section, we introduce the overview of \textsc{CapRecover}. During training, \textsc{CapRecover} aligns these intermediate image features with the corresponding ground truth captions/labels, effectively learning the mapping between visual representations and textual descriptions. During inference, \textsc{CapRecover} relies exclusively on the intermediate features to generate the image caption. While we describe our model primarily in the context of image caption reconstruction, the overall framework applies equally to image label recovery with only minor task-specific adaptations.

\subsection{Overview of \textsc{CapRecover}}
\label{subsec:overview_model}

\par\noindent As shown in Figure \ref{fig:model_overview}, \textsc{CapRecover} is composed of three primary modules: (1) Feature Projection Module, (2) Feature-Text Alignment Module, and (3) Caption Generation Module.

\begin{table}[t]
    \centering
    \caption{Information of different victim models and their intermediate layers' shapes.}
    \label{tab:victim_info}
    \vspace{-1.0em} 
    \begin{threeparttable}
        \sizefive
        \setlength{\tabcolsep}{4pt}
        \renewcommand{\arraystretch}{1.05}
        \begin{tabular}{ccc}
            \toprule
            \textbf{Victim model} & \textbf{Intermediate layer} & \textbf{Output feature dimension$^{*}$} \\
            \midrule
            $\text{CLIP}_{\text{ViT16}}$   & base & 512 \\
             $\text{CLIP}_{\text{ViT32}}$ & no-proj & 768 \\
             \hline
            \multirow{5}{*}{\makecell[c]{ResNet50\\(ResNet101)}} & base & 1024 (512) \\
             & layer1 & [256, 56, 56] $\rightarrow$ 1024 \\
             & layer2 & [512, 28, 28] $\rightarrow$ 1024 \\
             & layer3 & [1024, 14, 14] $\rightarrow$ 1024 \\
             & layer4 & [2048, 7, 7] $\rightarrow$ 1024 \\
             \hline
            MobileNetV2 & \multirow{2}{*}{base} & \multirow{2}{*}{1000} \\
            MobileNetV3 & & \\
            \bottomrule
        \end{tabular}
        \begin{tablenotes}
            \item \hspace{-1.2em}$^{*}$ {We use a ResNet-based projection module to transform those intermediate features retaining spatial dimensions (e.g., [32, 112, 112]) into a unified vectorized feature space (i.e., $\mathbb{R}^{1024}$).}
        \end{tablenotes}
    \end{threeparttable}
    \vspace{-1.2em}
\end{table}

\subsubsection{Feature Projection Module}
\label{subsec:caprecover_feature_projection}

\textsc{CapRecover} maps the victim model’s intermediate features into a dimensionally fixed (e.g., 1024) feature space via a projection layer. For example, given an input image $ I_i $, the victim model's middle layers produce intermediate features $ \mathbf{F}_i $. When these features are already in vector form, i.e., $\mathbf{F}_i \in \mathbb{R}^d $, we can simply apply a simple linear projection for $\mathbf{F}_{i}$, which is
\begin{equation}
\mathbf{F}_i^{proj} = \mathbf{W}_p \mathbf{F}_i + \mathbf{b}_p, \quad \mathbf{W}_p \in \mathbb{R}^{d' \times d}, \quad \mathbf{b}_p \in \mathbb{R}^{d'}
\end{equation}
where $ \mathbf{F}_i^{proj} \in \mathbb{R}^{d'} $ is the projected feature. $d^{'}$ is the dimension of the projected feature space. $\textbf{W}_p$ and $\mathbf{b}_p$ are learnable parameters of the feature projection layer.

In cases where the intermediate outputs retain spatial dimensions (i.e., $\mathbf{F}_i \in \mathbb{R}^{C \times H \times W}$ with $C$, $H$ and $W$ denoting the channel, height, and width, respectively), \textsc{CapRecover} first employs a ResNet-based projection module $g(\cdot)$ to convert these spatial features into a vectorized form. The transformation is then expressed as:
\begin{equation}
\mathbf{F}_i^{proj} = \mathbf{W}_p \cdot g(\mathbf{F}_i) +\mathbf{b}_p, \quad g: \mathbb{R}^{C \times H \times W} \to \mathbb{R}^{d'}
\end{equation}
where $\mathbf{W}_p$ and $\mathbf{b}_p$ are learnable parameters of  $g(\cdot)$. This additional projection module ensures that \textsc{CapRecover} can consistently process intermediate features from various victim models and different network layers by mapping them into a unified feature space.

\subsubsection{Feature-Text Alignment Module}
\label{subsec:caprecover_feature_text_alignment}

\textsc{CapRecover} employs an alignment module (for image features and captions) to establish a semantic correspondence between the projected intermediate image features and the ground truth caption. Specifically, while training our \textsc{CapRecover}, the alignment module first tokenizes and embeds the ground truth caption for each image, resulting in a sequence of text embeddings $\mathbf{T}_i$. Second, to fuse these textual cues with the visual information, \textsc{CapRecover} further employs a Q-Former model that leverages $K$ trainable query tokens $\mathbf{Q} \in \mathbb{R}^{K \times d'}$.

The Q-Former performs cross-modal attention by interacting with the projected features $\mathbf{F}_i^{proj}$ and the text embedding $\mathbf{T}_i$, producing enriched embeddings $ \mathbf{Z}_i $ that capture the alignment between visual and textual modalities, i.e.,
\begin{equation}
\mathbf{Z}_i = \text{Q-Former}(\mathbf{Q}, \mathbf{F}_i^{proj}, \mathbf{T}_i),
\end{equation}
where $\mathbf{Z}_i \in \mathbb{R}^{K \times d''}$ and $d''$ is the hidden size of the Q-Former. $\mathbf{Z}_i$ is further projected to match the input space of the language model:
\begin{equation}
\mathbf{E}_i = \mathbf{W}_l \mathbf{Z}_i, \quad \mathbf{W}_l \in \mathbb{R}^{d_{LM} \times d''}
\end{equation}
where $ \mathbf{E}_i \in \mathbb{R}^{K \times d_{LM}} $ serves as input to the language model.
At inference time, when the ground truth caption is not available, the Feature-Text Alignment module relies solely on the projected image features $\mathbf{F}_i^{proj}$ to generate the enriched embeddings $\mathbf{Z}_i$. These embeddings are subsequently forwarded to the caption generation module, completing the reconstruction pipeline.


\begin{table*}
    \centering
    \caption{Experimental results of \textsc{CapRecover} attacking different victim models on three datasets.}
    \label{tab:exp_res_coco2017}
    \vspace{-1.0em}
    \begin{threeparttable}
        \sizefive
        \setlength{\tabcolsep}{4pt}
        \renewcommand{\arraystretch}{1.05}
        \begin{tabular}{ccccccccccc}
            \toprule
            \textbf{Dataset} & \textbf{Victim model}$^{*}$ & \textbf{BLEU-1} & \textbf{BLEU-2} & \textbf{BLEU-3} & \textbf{BLEU-4} & \textbf{METEOR} & \textbf{ROUGE\_L} & \textbf{CIDEr} & \textbf{SPICE} & \makecell[c]{\textbf{Cosine Similarity (\%)}$^{\ddagger}$} \\
            \midrule
            \multirow{6}{*}{COCO2017} & $\text{CLIP}_{\text{ViT16}}$    & 0.72        & 0.55        & 0.41  & 0.30 & 0.26 & 0.53 & 0.99 & 0.19 & 84.38       \\
            & $\text{CLIP}_{\text{ViT32}}$     & 0.70        & 0.53        & 0.39  & 0.29 & 0.26 & 0.53 & 0.95 & 0.19 & 80.38       \\
            & RN50     & 0.70        & 0.52        & 0.38  & 0.28 & 0.25 & 0.52 & 0.90 & 0.18 & 76.84       \\
            & RN101     & 0.70        & 0.52        & 0.39  & 0.28 & 0.25 & 0.53 & 0.93 & 0.18 & 79.98       \\
            & MNV2     & 0.39        & 0.18        & 0.10  & 0.06 & 0.11 & 0.31 & 0.09 & 0.03 & ~0.44     \\
            & MNV3     & 0.40        & 0.19        & 0.10  & 0.08 & 0.11 & 0.31 & 0.10 & 0.03 & ~2.74       \\
            \hline
            \hline
            \multirow{6}{*}{Flickr8K} & $\text{CLIP}_{\text{ViT16}}$    & 0.30        & 0.15        & 0.08  & 0.05 & 0.13 & 0.27 & 0.54 & 0.19 & 22.40         \\
            & $\text{CLIP}_{\text{ViT32}}$     & 0.29        & 0.15        & 0.08  & 0.05 & 0.12 & 0.26 & 0.48 & 0.17 & 18.40        \\
            & RN50     & 0.28        & 0.14        & 0.08  & 0.04 & 0.12 & 0.25 & 0.46 & 0.17 & 16.50        \\
            & RN101    & 0.28        & 0.14        & 0.08  & 0.05 & 0.12 & 0.25 & 0.47 & 0.17 & 18.00       \\
            & MNV2     & 0.20        & 0.08        & 0.04  & 0.02 & 0.07 & 0.17 & 0.16 & 0.06 & ~1.20        \\
            & MNV3     & 0.21        & 0.08        & 0.04  & 0.02 & 0.07 & 0.17 & 0.17 & 0.06 & ~1.80        \\
            \hline
            \hline
            \multirow{6}{*}{ImageNet-1K} & $\text{CLIP}_{\text{ViT16}}$     & 0.45        & 0.30        & 0.21  & 0.15 & 0.18 & 0.41 & 1.18 & 0.2 & 40.78       \\
            & $\text{CLIP}_{\text{ViT32}}$     & 0.44        & 0.29        & 0.20  & 0.14 & 0.18 & 0.40 & 1.08 & 0.19 & 36.11        \\
            & RN50     & 0.42        & 0.27        & 0.18  & 0.13 & 0.16 & 0.38 & 0.93 & 0.16 & 27.00       \\
            & RN101     & 0.42        & 0.27        & 0.18  & 0.13 & 0.16 & 0.38 & 0.95 & 0.16 & 27.78      \\
            & MNV2     & 0.29        & 0.15        & 0.07  & 0.03 & 0.09 & 0.26 & 0.18 & 0.03 & ~1.67      \\
            & MNV3     & 0.29        & 0.14        & 0.07  & 0.04 & 0.09 & 0.26 & 0.22 & 0.04 & ~6.11    \\
            \bottomrule
        \end{tabular}
        \begin{tablenotes}
            \item \hspace{-1.2em}$^{*}$ {``RN50'' (``RN101'') denotes ResNet50 (ResNet101) and ``MNV2'' (MNV3) denotes MobileNetV2 (MobileNetV3). }
            \item \hspace{-1.2em}$^{\ddagger}$ {We calculate the proportion of cosine similarities that are greater than a predefined threshold, which is empirically set to 0.7 in our paper.}
        \end{tablenotes}
    \end{threeparttable}
    \vspace{-1.1em}
\end{table*}

\subsubsection{Caption Generation}
\label{subsec:caprecover_caption_generation}

By using these outputs from the Feature-Text Alignment Module, \textsc{CapRecover} further employs a large language model (LLM) to interpret the compressed image representations and generate the final caption that accurately describes the image's semantic content. Specifically, the LLM processes the input embeddings $\mathbf{E}_i$ and, if provided, extra text input embeddings $\mathbf{T}_i$ (e.g., prompts), to generate the caption $C_i$ for the image $I_i$. This caption generation process is modeled autoregressively as follows:
\begin{equation}
P(C_i \mid \mathbf{E}_i, \mathbf{T}_i) = \prod_{t=1}^{T} P(c_{i,t} \mid c_{i,<t}, \mathbf{E}_i, \mathbf{T}_i),
\end{equation}
where $c_{i,t}$ denotes the token generated at time step $t$ and $c_{i,<t}$ represents all preceding tokens before time step $t$.

\subsection{Model Training Objective and Settings}
\label{subsec:model_obj}

\par\noindent To train the model, we minimize the cross-entropy loss between the generated caption $ C_i $ and the ground-truth caption $ C_i^{*} $:
\begin{equation}
\mathcal{L} = - \frac{1}{N} \sum_{i=1}^N \sum_{t=1}^T \log P(c_{i,t}^{*} \mid c_{i,<t}^{*}, \mathbf{E}_i, \mathbf{T}_i)
\end{equation}
where $ N $ is the batch size, $ T $ is the caption length, and $ c_{i,t}^{*} $ is the ground-truth.

\label{subsec:model_setting}



The feature projection module in \textsc{CapRecover} is initialized with a random distribution, while employing a pre-trained Q-Former model for the feature-text alignment and a pre-trained OPT model for language generation. To focus the training on aligning the visual features with the corresponding textual information, we freeze the parameters of the language model and update only those in the feature projection and feature-text alignment modules.

\textsc{CapRecover} is trained for six epochs with a learning rate of $5e-5$. The training batch size is configured to 16, while the testing batch size is set at 8. All experiments are conducted on a cloud server equipped with a single NVIDIA RTX 4090 (24 GB memory).

\section{Experiments on Caption Reconstruction}
\label{sec:experiments}

\subsection{Experimental Settings}
\label{subsec:settings}

\subsubsection{Datasets}
\label{subsec:exp_dst}

\begin{figure*}[t]
    \hspace{-1.5em}
    \begin{tabular}{cc}
        
        \subfigure[{Evaluation on COCO2017 dataset.}]{
            \begin{minipage}[t]{0.31\textwidth}
                \centering
                \label{fig:bleu1_on_coco2017}
                \includegraphics[width=0.89\textwidth]{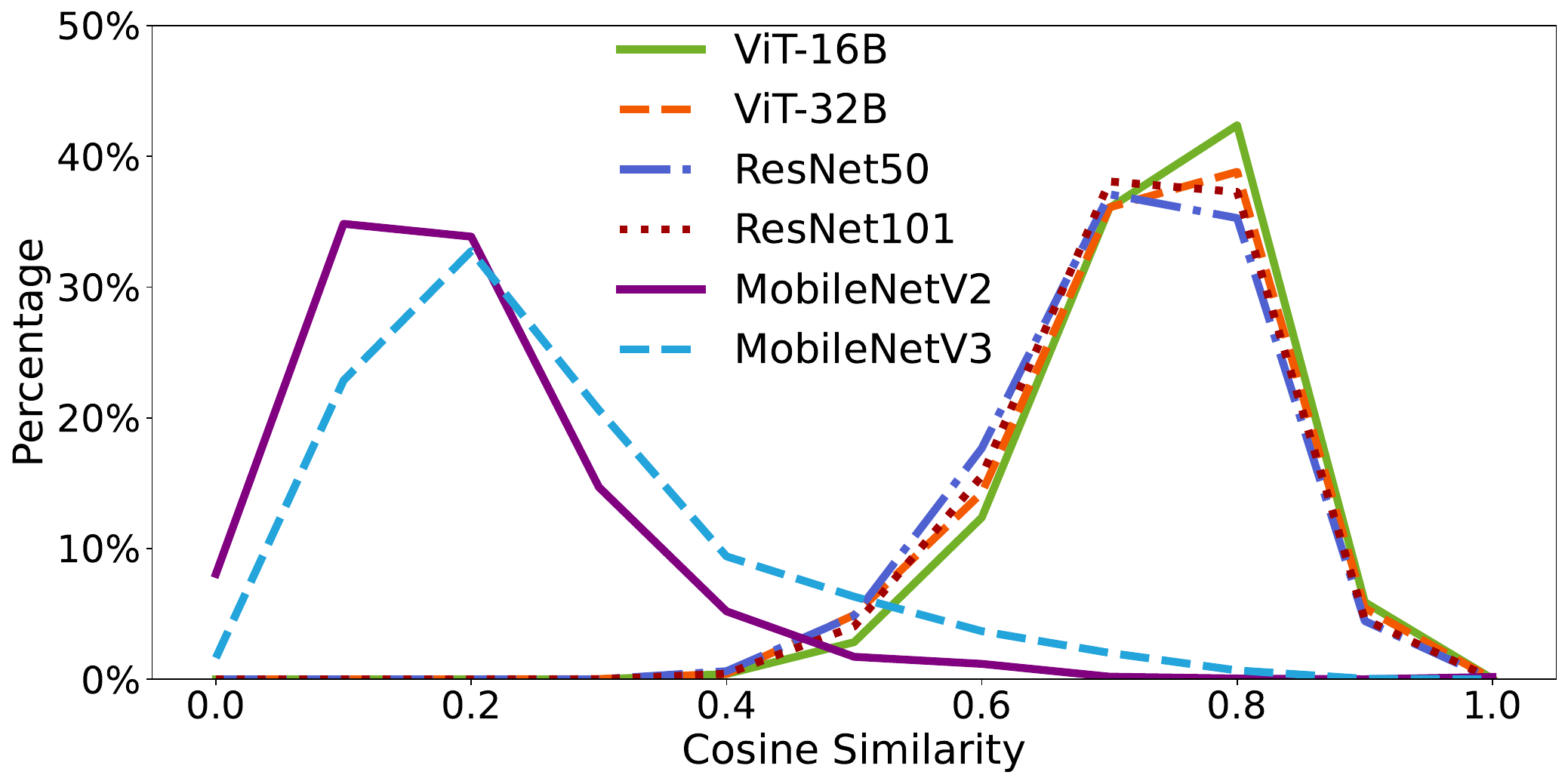}
        \end{minipage}}

        \subfigure[{Evaluation on Flickr8K dataset.}]{
            \begin{minipage}[t]{0.31\textwidth}
                \centering
                \label{fig:bleu1_on_flickr8k}
                \includegraphics[width=0.89\textwidth]{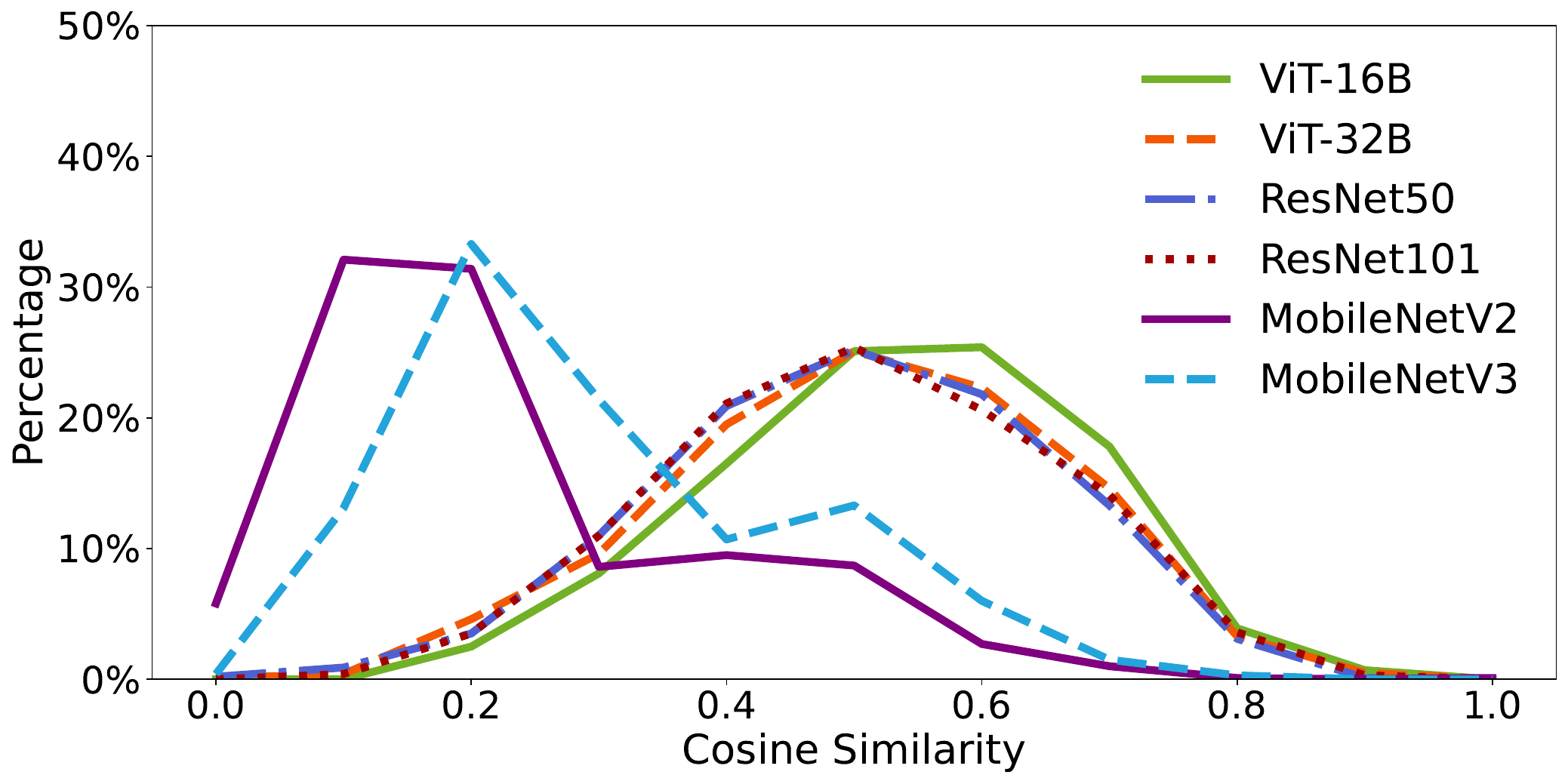}
        \end{minipage}}

        \subfigure[{Evaluation on ImageNet-1K dataset.}]{
            \begin{minipage}[t]{0.31\textwidth}
                \centering
                \label{fig:bleu1_on_imagenet1k}
                \includegraphics[width=0.89\textwidth]{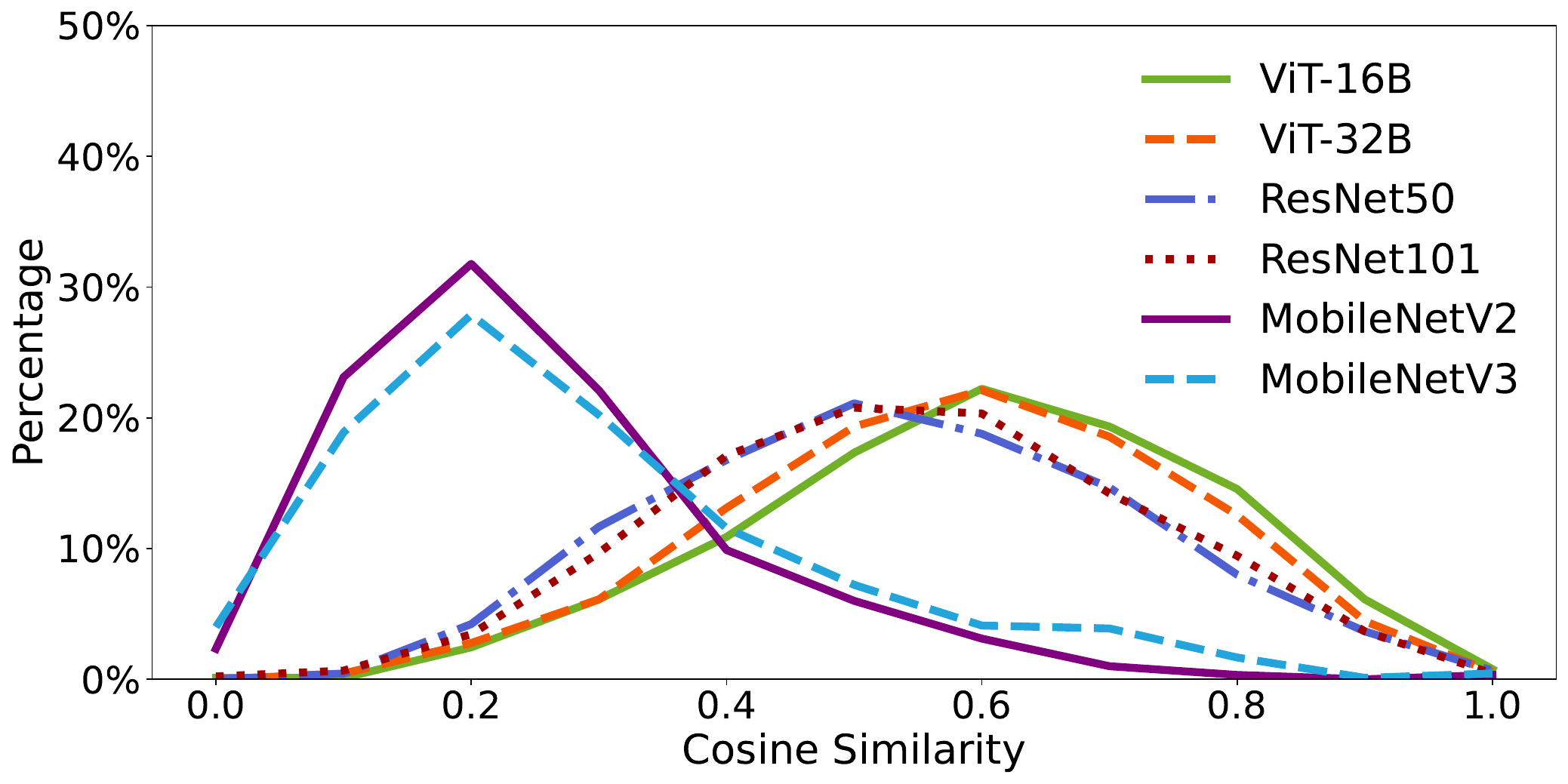}
        \end{minipage}
        }
    \end{tabular}
    \vspace{-1.3em}

    \caption{
        Distribution of cosine similarities across three datasets. We use intermediate features extracted from the final layer of the victim model to train \textsc{CapRecover}. We analyze how other intermediate layers' features impact performance in Sec.~\ref{subsec:further_study_analysis}.
    }
	\label{fig:cosine_similarity_on_three_dst}
    \vspace{-1.0em}
\end{figure*}

To comprehensively evaluate the effectiveness of \textsc{CapRecover}, we adopt three widely-used datasets: COCO2017 \cite{lin2014microsoft}, Flickr8K \cite{dst_flickr8k}, and ImageNet-1K \cite{dst_imagenet}. For the ImageNet-1K dataset, we use Qwen2.5 \cite{yang2024qwen2} to generate captions for the images. We employ generated captions for ImageNet-1K due to: (1) the original ImageNet-1K dataset does not provide human-annotated captions, and (2) recent research \cite{nguyen2023improving, lei2023image} demonstrates the effectiveness and semantic accuracy of captions generated by advanced VLMs. We will clarify this in the revised version. Given the large size of the original ImageNet-1K dataset, we randomly sample 12,000 images for the training set and 1,000 images for the test set. More details about these datasets are provided in Table \ref{tab:dsts_used}.









\subsubsection{Victim Models}
\label{subsec:model_layer}
We focus on three widely adopted visual models commonly utilized in Vision-Language Models (VLMs) and deployed on edge devices: Vision Transformer \cite{dosovitskiy2020image} (ViT), ResNet \cite{he2016deep}, and MobileNet (MobileNetV2 \cite{sandler2018mobilenetv2} and MobileNetV3 \cite{howard2019searching}). In practice, ResNet (e.g., ResNet50 and ResNet101) serves as the image encoder in VLMs like CLIP \cite{clip} and UPL \cite{huang2022unsupervised}. ViT (e.g., ViT-16B and ViT-32B) serves as the visual module in VLMs such as CLIP \cite{clip} and LlaVa \cite{liu2023llava}. As a lightweight convolutional neural network optimized for mobile applications, MobileNet is widely deployed on edge devices like mobile phones, offering efficient performance for on-device inference.

We analyze both the final output of the victim model (referred to as the ``base'' output) and the intermediate output before the final linear projection layer (denoted as ``no-proj'' for ViT-based models and ``layer4'' for ResNet-based models). Additionally, we examine the impact of different middle layers within the victim models (e.g., ``layer1''$\thicksim$``layer4'' in ResNet50) on reconstruction performance.





\subsubsection{Evaluation Metrics}
\label{subsec:metrics}



We evaluate \textsc{CapRecover}'s performance using two main categories of metrics: standard metrics and semantic similarity metrics based on cosine similarity.

Common Metrics:
We adopt widely used evaluation measures, including: BLEU-1$\thicksim$BLEU-4, METEOR, ROUGE\_L, CIDEr, and SPICE, to assess the quality of the generated captions. These metrics quantify how closely the generated captions match the ground truth captions in terms of lexical overlap. We primarily rely on ROUGE-L as our main metric, since it captures structural alignment and semantic completeness more effectively than n-gram-based scores. We consider ROUGE-L scores above 0.3 as indicative of moderate attack success, reflecting partial semantic recovery, and scores above 0.5 as indicative of successful attacks that capture most of the key semantic content.

Embedding-Based Cosine Similarity:
In addition to the common metrics, we use a pre-trained embedding model \cite{yang2024qwen2} to project both the generated and ground truth captions into a shared semantic space. We then compute the cosine similarity between these embeddings to measure the semantic alignment between the captions.  We interpret similarity values above 0.7 as successful attacks.

Note that as far as we know, our work mainly focuses on the direct recovery of image captions or labels from leaked intermediate features rather than reconstructing images first. To our knowledge, no prior studies address this specific problem.

\subsection{Experimental Results}
\label{subsec:exp_res}

        




\subsubsection{Overall results}

We evaluate the performance of \textsc{CapRecover} on six victim models across three benchmark datasets: COCO2017, Flickr8K, and ImageNet-1K. As shown in Table~\ref{tab:exp_res_coco2017}, \textsc{CapRecover} achieves the strongest results on the COCO2017 dataset. For example, when attacking $\text{CLIP}_{\text{ViT16}}$, the model achieves a BLEU-1 score of 0.72 and a ROUGE-L score of 0.53, with 84.38\% of generated captions exceeding a cosine similarity threshold of 0.7—indicating strong semantic and structural alignment with the ground truth. Similar performance is observed for other ViT- and ResNet-based victim models, with ROUGE-L scores consistently around 0.52–0.53, which we consider indicative of successful semantic inversion.


\begin{figure*}[t]
    \centering
    \includegraphics[width = 0.91\linewidth]{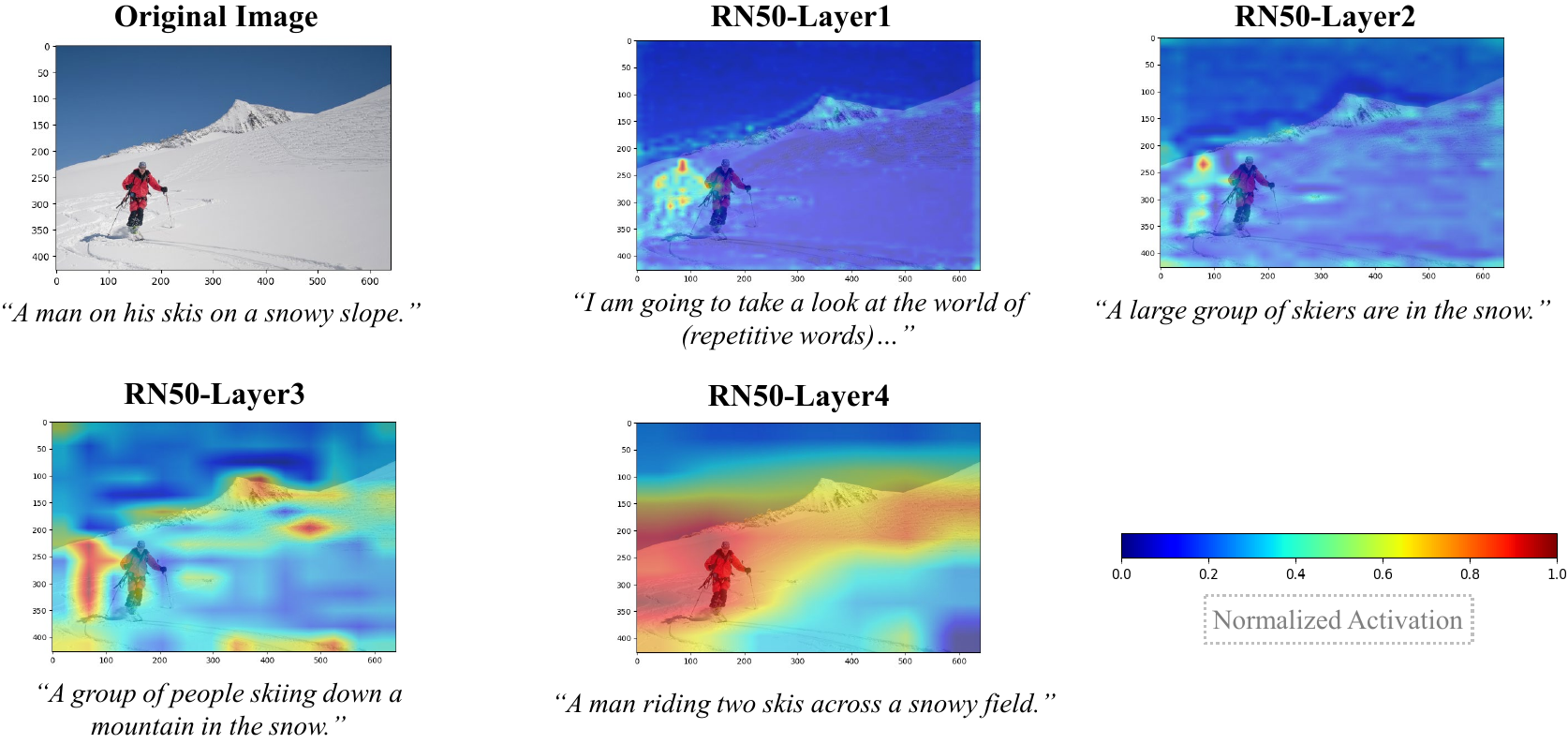}
    \vspace{-0.8em}
    \caption{Example of visualizing the heatmaps of ResNet50's different middle layers. Below each figure is the generated/ground truth caption. These figures demonstrate that the shallow layer (e.g., RN50-Layer1) pays more attention to edges and local features, while the deeper the layer (e.g., RN50-Layer4), the more attention is paid to the more semantic areas in the image.}
    \label{fig:rn50_heatmap}
    \vspace{-1.0em}
\end{figure*}

\begin{figure*}[t]
    \hspace{-1.3em}
    \begin{tabular}{ccccc}
    
        \subfigure[{ResNet50-layer1.}]{
            \begin{minipage}[t]{0.19\textwidth}
                \centering
                \label{fig:bleu1_on_rn50_layer1}
                \includegraphics[width=0.97\textwidth]{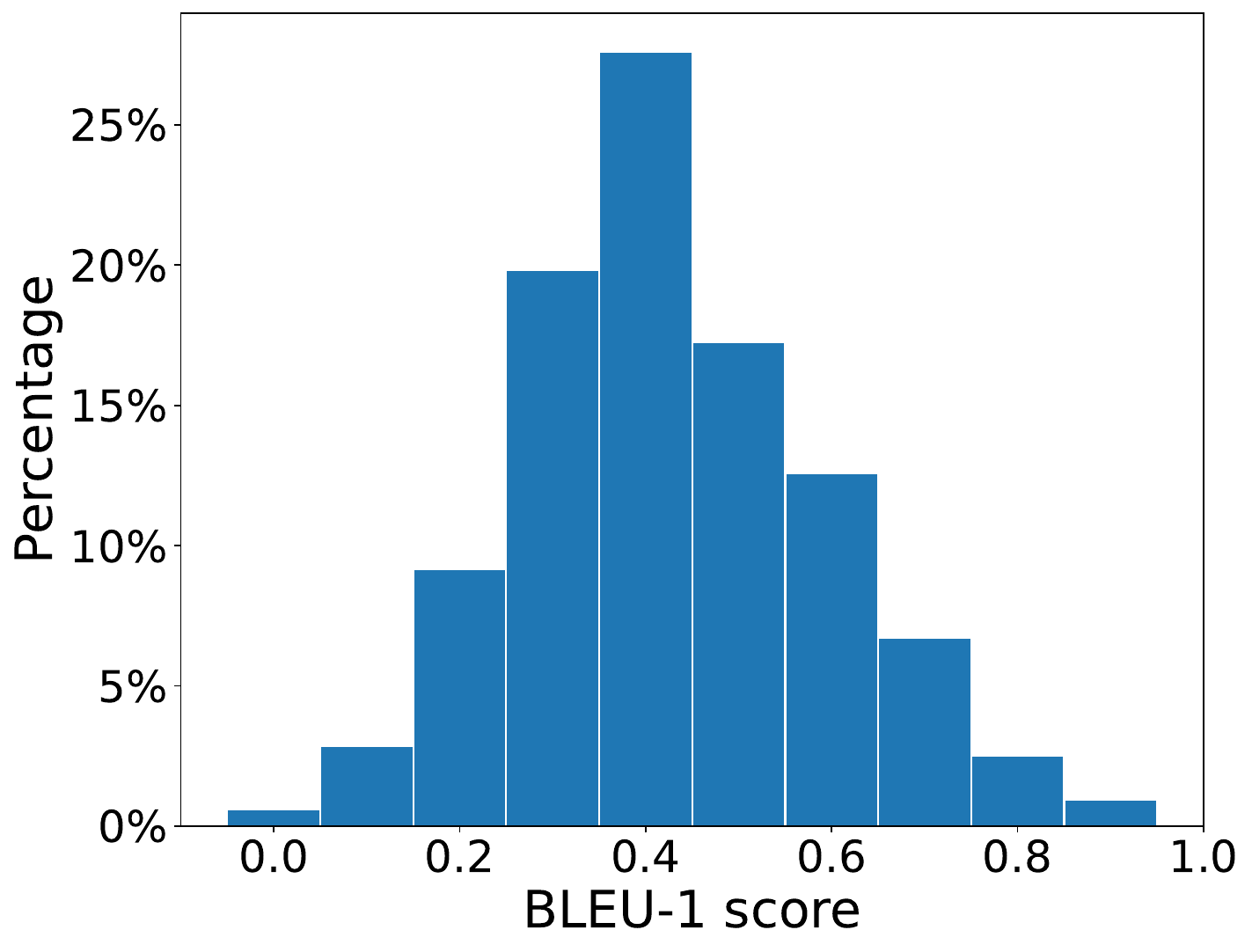}
        \end{minipage}}

        \subfigure[{ResNet50-layer2.}]{
            \begin{minipage}[t]{0.19\textwidth}
                \centering
                \label{fig:bleu1_on_rn50_layer2}
                \includegraphics[width=0.97\textwidth]{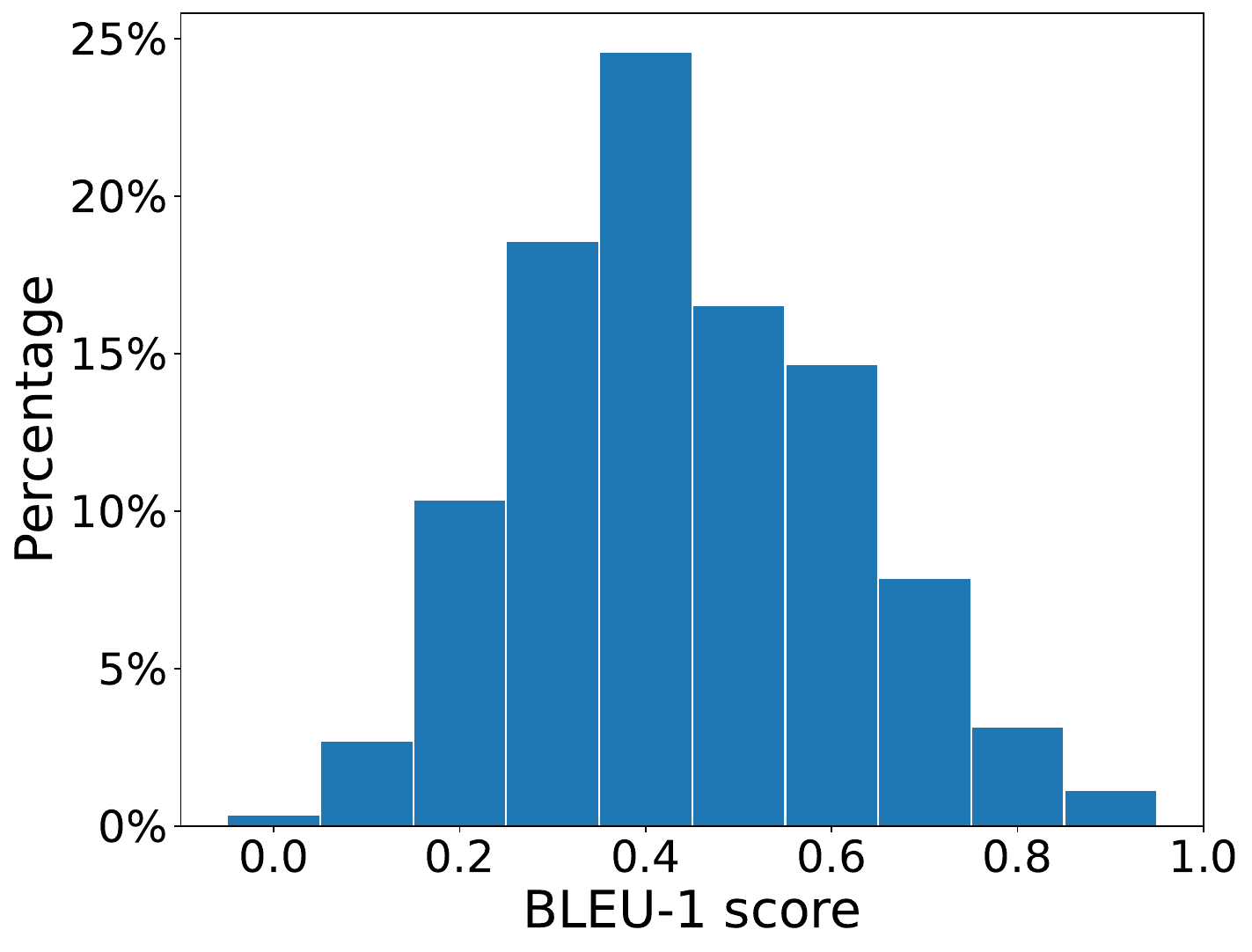}
        \end{minipage}}

        \subfigure[{ResNet50-layer3.}]{
            \begin{minipage}[t]{0.19\textwidth}
                \centering
                \label{fig:bleu1_on_rn50_layer3}
                \includegraphics[width=0.97\textwidth]{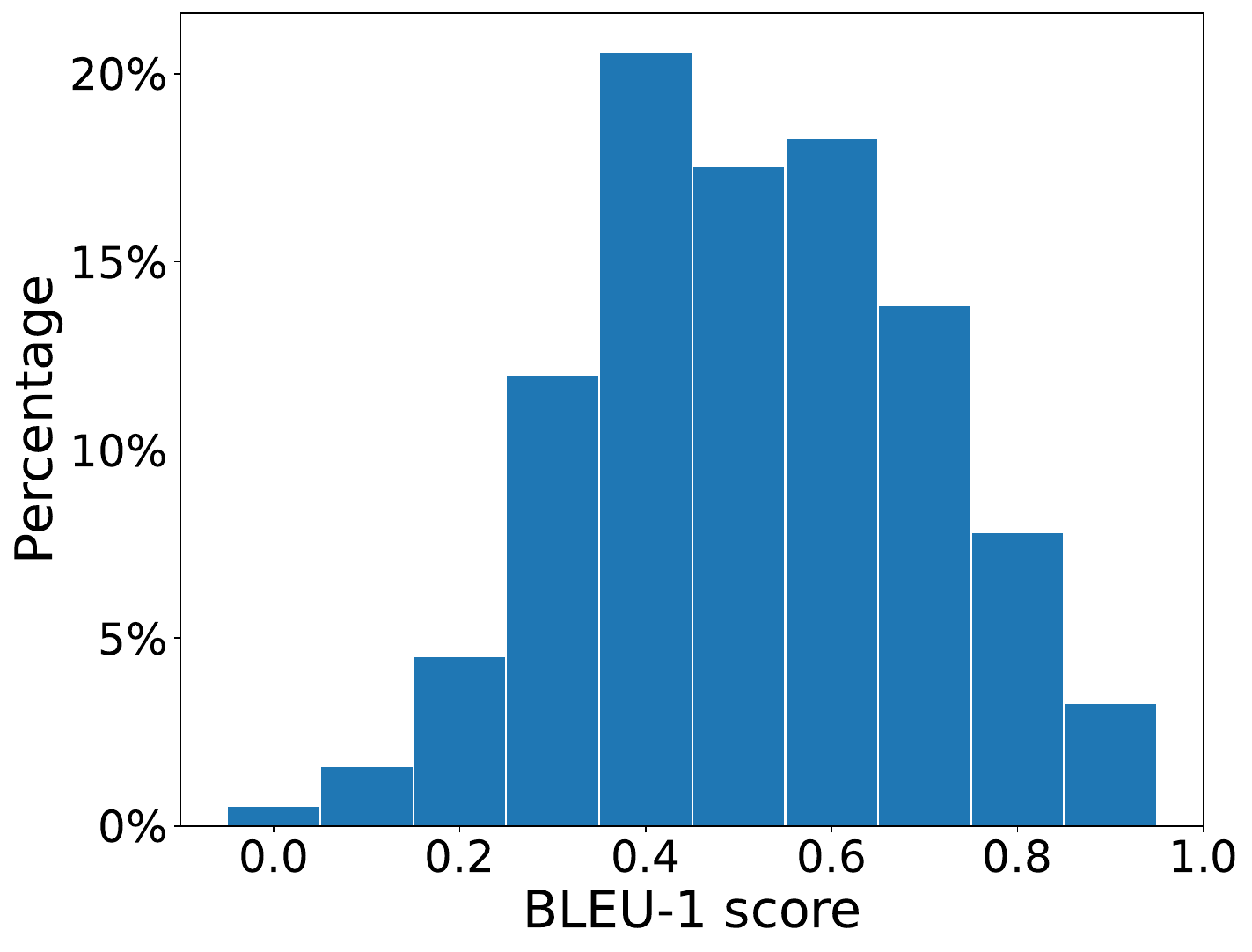}
        \end{minipage}}

        \subfigure[{ResNet50-layer4.}]{
            \begin{minipage}[t]{0.19\textwidth}
                \centering
                \label{fig:bleu1_on_rn50_layer4}
                \includegraphics[width=0.97\textwidth]{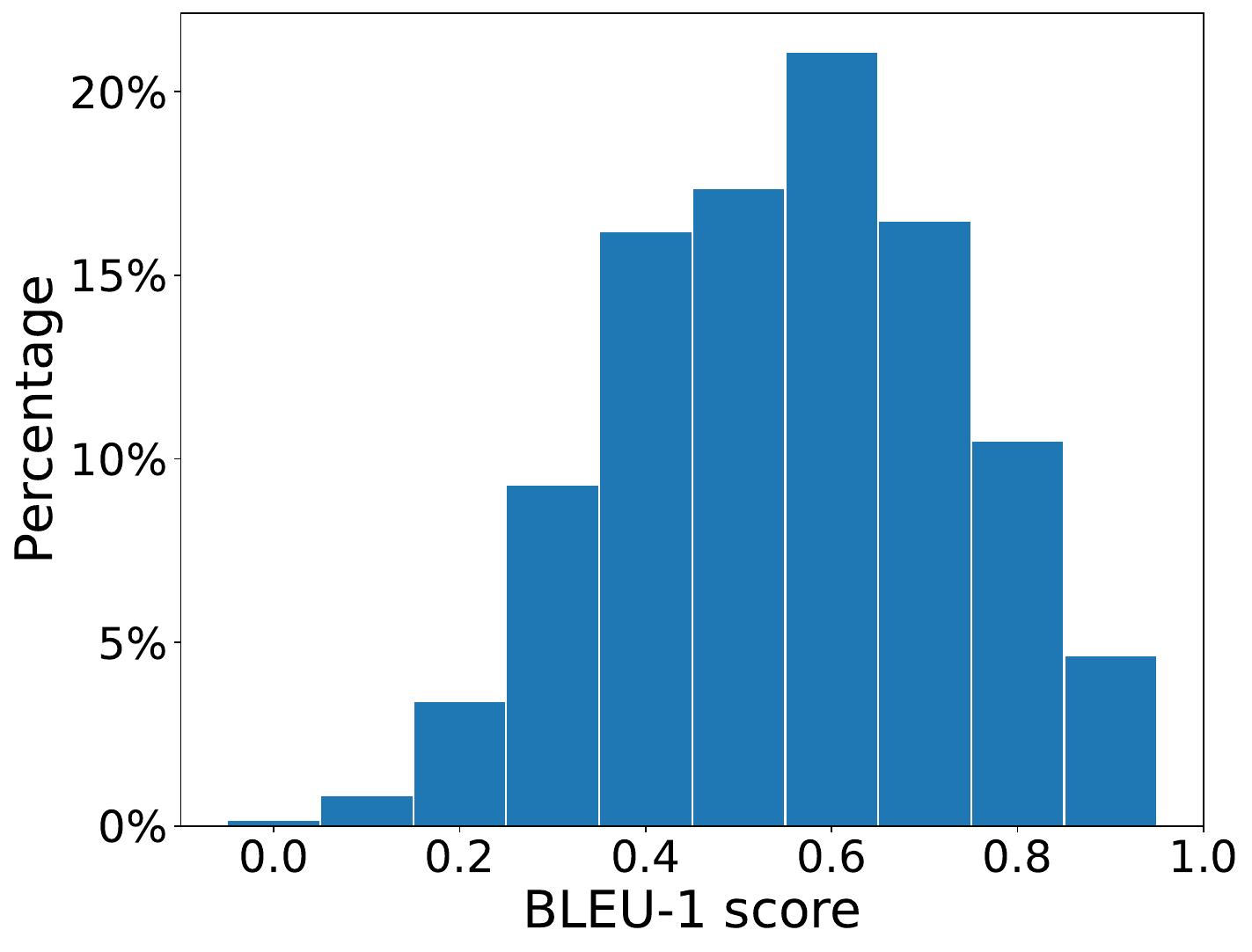}
        \end{minipage}}

        \subfigure[{ResNet50-base.}]{
            \begin{minipage}[t]{0.19\textwidth}
                \centering
                \label{fig:bleu1_on_rn50_base}
                \includegraphics[width=0.97\textwidth]{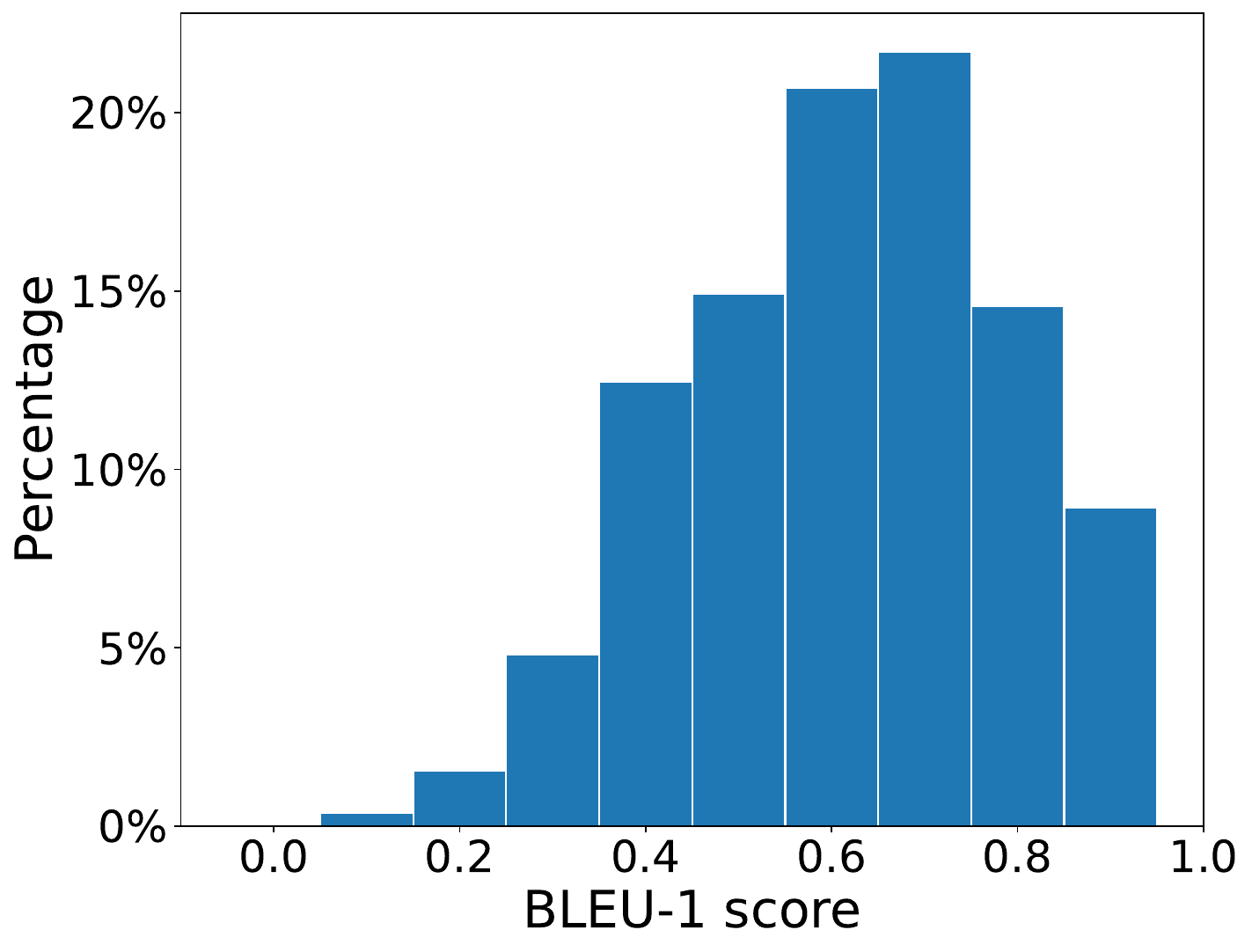}
        \end{minipage}}
        
    \end{tabular}
    \vspace{-1.1em}

    \caption{
        Evaluate the distributions of cosine similarity on the COCO2017 dataset. We train \textsc{CapRecover} using the intermediate image features produced by their final linear projection layers. We further discuss how other middle layers' intermediate features affect \textsc{CapRecover}'s performance in Sec. \ref{subsec:further_study_analysis}.
    }
	\label{fig:bleu1_on_rn50_coco}
    \vspace{-0.8em}
\end{figure*}

In contrast, performance on the Flickr8K dataset is substantially lower across all models. For instance, CLIP\textsubscript{ViT16} yields a BLEU-1 score of only 0.30 and a ROUGE-L of 0.27, and the proportion of cosine similarities exceeding 0.7 drops to 22.40\%. This degradation is largely due to the small size of Flickr8K (8,000 images) and the nature of its captions, which are shorter, less diverse, and often semantically sparse. Such properties limit the model’s ability to learn rich visual-to-text mappings and result in lower alignment on both lexical and structural metrics.

Results on ImageNet-1K fall between the two extremes. Despite being a classification dataset without explicit captions, ViT- and ResNet-based models still enable moderate recovery: CLIP\textsubscript{ViT16} yields a BLEU-1 of 0.45, ROUGE-L of 0.41, and cosine similarity over 40\%. These results suggest that classification-pretrained encoders implicitly retain a significant amount of caption-relevant semantic information in their intermediate features.

Across all three datasets, MobileNetV2 and MobileNetV3 consistently show the weakest performance. For example, on COCO2017, their ROUGE-L scores are only 0.31 and their cosine similarities barely exceed 3\%, indicating poor semantic preservation. We attribute this to the lightweight, efficiency-oriented design of MobileNet models. MobileNet utilizes depthwise separable convolutions and aggressive dimensionality reduction strategies designed for efficiency, which reduce model complexity but significantly compromise the model’s ability to capture detailed and high-level semantic features \cite{li2022efficientformer,li2023rethinking,vasu2023fastvit}. This architectural limitation inherently leads to less detailed intermediate features, explaining the lower effectiveness of CapRecover on MobileNet. We will clarify this in the revised version.
A more detailed analysis of layer-wise performance is provided in Section~\ref{subsec:further_study_analysis}.

\vspace{-0.5em}
\subsection{Further Study on Middle Layers}
\label{subsec:further_study_analysis}

\par\noindent As shown in Table \ref{tab:exp_rn50_mid_layer}, we employ $\text{CLIP}_{\text{ViT16}}$ and ResNet50 as victim models to investigate how intermediate image features from different layers impact caption reconstruction. Our analysis reveals that features extracted from shallow layers contribute minimally to caption reconstruction because they primarily capture low-level visual characteristics, such as edges and textures. 

In contrast, the intermediate image features from deeper layers significantly enhance \textsc{CapRecover}’s performance, as evidenced by higher BLEU-1 scores compared to those obtained from shallow layers. This finding suggests that as convolutional layers deepen, they capture more specific and meaningful semantic information from the image. Figure \ref{fig:bleu1_on_rn50_coco} illustrates this trend by showing that the BLEU-1 score distribution for captions generated by \textsc{CapRecover} on the COCO2017 dataset shifts to the right as layer depth increases, reflecting improved overall prediction accuracy.

Furthermore, Figure \ref{fig:rn50_heatmap} visualizes heat maps of different convolutional layers in ResNet50 for a target image. When \textsc{CapRecover} uses features from a shallow layer (e.g., ResNet50-layer1), which captures basic semantics such as the edges of a mountain or human, the generated captions are less accurate and may even meaningless. However, as the intermediate features come from deeper layers, \textsc{CapRecover} gradually captures more relevant information from the image (such as ``snow'', ``skiing'' and ``man''). When utilizing features from a deep middle layer (e.g., ResNet50-layer4), the generated caption closely approximates the ground truth caption.


\section{Experiments on Label Recovery}
\label{sec:exp_img_classification}

In this section, we explore extending \textsc{CapRecover} to additional Vision-Language Model (VLM) application scenarios—specifically, \textit{image label recovery}. While our method is primarily introduced in the context of image caption reconstruction, the underlying architecture and attack strategy are general and easily transferable. To adapt \textsc{CapRecover} (as shown in Figure \ref{fig:model_overview}) for classification reconstruction tasks, we replace the original large language model (LLM) component with a standard linear classifier, as our preliminary experiments indicate that employing a simple classifier is enough to achieve high Top-1 accuracy.

\subsection{Experimental Settings}
\label{subsec:exp_set_img_classification}


\subsubsection{Dataset} As shown in Table \ref{tab:dsts_used}, we evaluate \textsc{CapRecover} on the CIFAR-10 and TinyImageNet datasets. CIFAR-10 consists of 60,000 images evenly distributed across 10 classes, with 50,000 images used for training and 10,000 for testing. TinyImageNet contains 200 classes with 500 training images and 50 validation images per class (a total of 100,000 images). Due to the increased dataset complexity and class granularity, TinyImageNet serves as a challenging benchmark for label recovery attacks.

\subsubsection{Victim models} Similar to the setting in Sec. \ref{subsec:model_layer}, the victim models we selected are ResNet-50 (i.e., RN50) and a CLIP variant with a ViT-B/32 backbone (i.e., $\text{CLIP}_{\text{ViT16}}$), from which intermediate visual features are extracted as inputs to our \textsc{CapRecover}. We use the final output of these victim models (i.e., ``base'' output) as the intermediate features.

\subsubsection{Model settings} We train the label recovery model using the Adam optimizer with a learning rate of $5 \times 10^{-4}$, a batch size of 64 for training and 16 for evaluation. The model is trained for 5 epochs in total. All experiments are conducted using standard cross-entropy loss for classification.

\begin{table}[t]
    \centering
    \caption{Comparison of experimental results on ResNet50 and $\text{CLIP}_{\text{ViT16}}$ using different middle layers.}
    \label{tab:exp_rn50_mid_layer}
    \vspace{-1.0em}
    \begin{threeparttable}
        \sizefive
        \setlength{\tabcolsep}{4pt}
        \renewcommand{\arraystretch}{1.05}
        \begin{tabular}{ccccc}
            \toprule
            Victim model & Middle layer & BLEU-1 & CIDEr & Cosine Similarity (\%) \\
            \midrule
            \multirow{5}{*}{ResNet50} & layer1       & 0.24        & 0.19        & ~0.00         \\
             & layer2       & 0.51        & 0.31        & 43.76          \\
             & layer3       & 0.58        & 0.55        & 31.42          \\
             & layer4       & 0.62        & 0.68        & 85.64          \\
             & base         & 0.70        & 0.90        & 90.52       \\
             \hline
             \hline
             \multirow{2}{*}{$\text{CLIP}_{\text{ViT16}}$} & no-proj & 0.69 & 0.90 & 93.04 \\
              & base & 0.72 & 0.99 & 94.76 \\
            \bottomrule
        \end{tabular}
    \end{threeparttable}
\end{table}

\begin{figure}[t]
    \centering
    \includegraphics[width = 0.85\linewidth]{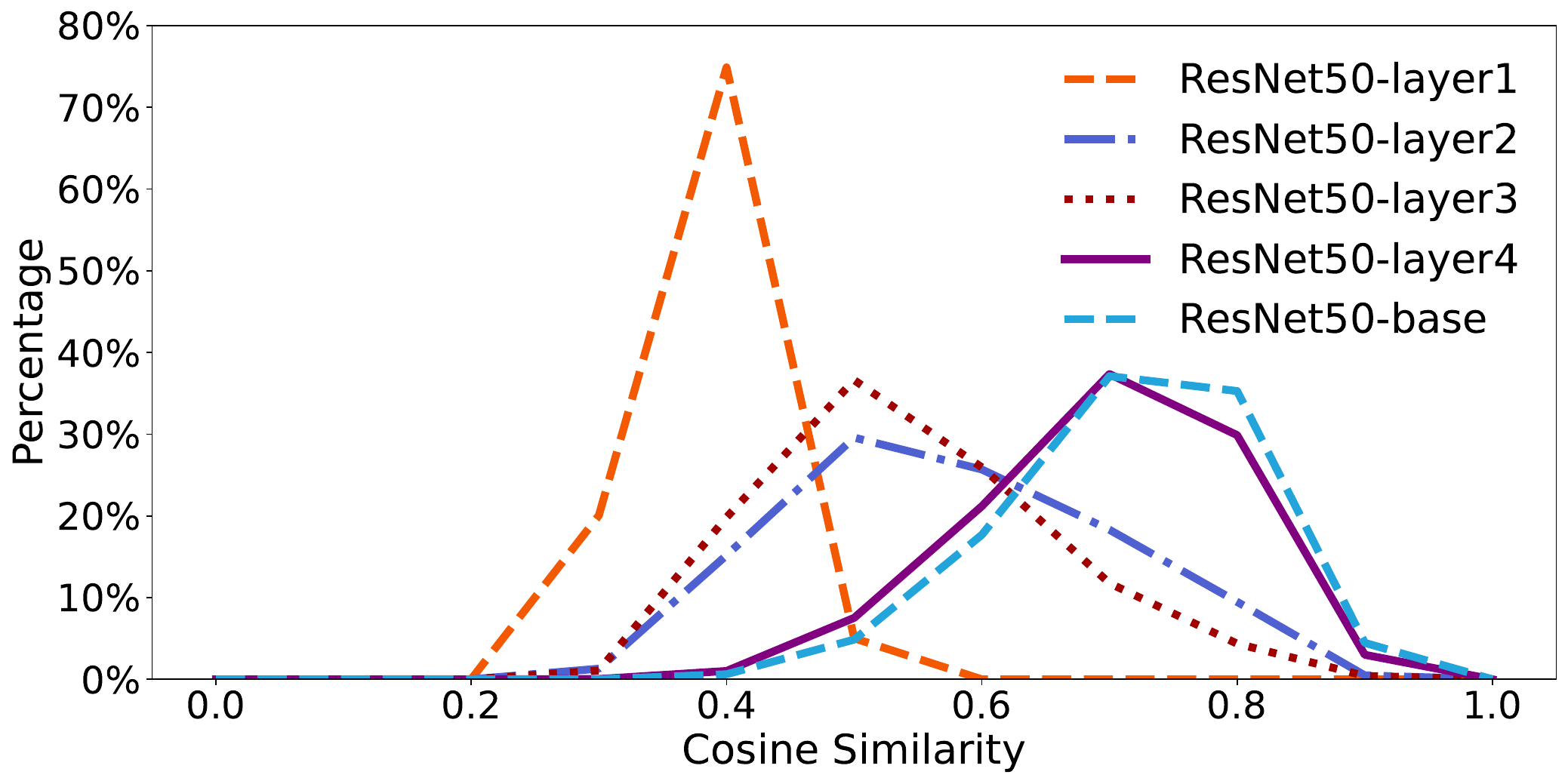}
    \vspace{-0.8em}
    \caption{Embedding Cosine Similarity distributions for different middle layers of ResNet50. The results indicate that \textsc{CapRecover} employs the deep layers and may perform better compared to the shallow layers.}
    \label{fig:exp_resnet50_cosine_similarity_middle_layers}
    \vspace{-0.5em}
\end{figure}

\subsection{Overall Experimental Results}
\label{subsec:exp_res_img_classification}

Table \ref{tab:exp_overall_res_on_dsts} shows the consistently strong performance of \textsc{CapRecover} on image label recovery tasks across two widely-used datasets (CIFAR-10 and TinyImageNet) and two victim models (ResNet50 and CLIP\textsubscript{ViT32}). Specifically, \textsc{CapRecover} achieves particularly high Top-1 and Top-5 accuracy scores on CIFAR-10 dataset. Notably, when attacking CLIP\textsubscript{ViT32}, \textsc{CapRecover} achieves a Top-1 accuracy of 92.71\% and Top-5 accuracy of 99.82\%, indicating near-perfect recovery of class labels.

While testing the TinyImageNet dataset, due to its larger number of classes and higher semantic complexity, \textsc{CapRecover} achieves lower accuracy scores compared to the results on CIFAR-10. However, even under this challenging setting, \textsc{CapRecover} still achieves 72.62\% Top-1 accuracy and 91.60\% Top-5 accuracy when targeting CLIP\textsubscript{ViT32}, demonstrating the model’s strong generalization ability across both datasets and victim architectures. In comparison, attacks on ResNet50 yield lower performance across both datasets, suggesting that visual representations from CLIP-based encoders are more vulnerable to semantic leakage.

\begin{table}[t]
    \centering
    \caption{Experimental results of image label recovery on CIFAR-10 and TinyImageNet datasets.}
    \label{tab:exp_overall_res_on_dsts}
    \vspace{-1.0em}
    \begin{threeparttable}
        \sizefive
        \setlength{\tabcolsep}{4pt}
        \renewcommand{\arraystretch}{1.10}
        \begin{tabular}{cccc}
            \toprule
            Victim model & datasets & Top-1 Accuracy (\%) & Top-5 Accuracy (\%) \\
            \midrule
            \multirow{2}{*}{ResNet50} & CIFAR-10       & 83.35   & 99.55    \\
             & TinyImageNet       & 60.13       & 83.79  \\
             \hline
             \multirow{2}{*}{$\text{CLIP}_{\text{ViT32}}$} & CIFAR-10  & 92.71   & 99.82 \\
              & TinyImageNet & 72.62 & 91.60  \\
            \bottomrule
        \end{tabular}
    \end{threeparttable}
    \vspace{-1.0em}
\end{table}

\begin{table}[t]
    \centering
    \caption{Results of image label recovery on CIFAR-10.}
    \label{tab:exp_img_classification}
    \vspace{-1.0em}
    \begin{threeparttable}
        \sizefive
        \setlength{\tabcolsep}{4pt}
        \renewcommand{\arraystretch}{1.05}
        \begin{tabular}{cccc}
            \toprule
            Class & Precision & Recall & F1-Score  \\
            \midrule
            Airplane    & 0.93   & 0.96  & 0.95 \\
            Automobile  & 0.96   & 0.97  & 0.97 \\
            Bird        & 0.91   & 0.90  & 0.91 \\
            Cat         & 0.85   & 0.85  & 0.85 \\
            Deer        & 0.91   & 0.93  & 0.92 \\
            Dog         & 0.88   & 0.87  & 0.87 \\
            Frog        & 0.92   & 0.94  & 0.93 \\
            Horse       & 0.97   & 0.94  & 0.96 \\
            Ship        & 0.96   & 0.96  & 0.96 \\
            Truck       & 0.97   & 0.96  & 0.96 \\
            \bottomrule
        \end{tabular}
    \end{threeparttable}
    \vspace{-1.0em}
\end{table}

\subsection{Further Analysis on CIFAR-10}
\begin{figure}[t]
    \centering
    \includegraphics[width = 0.90\linewidth]{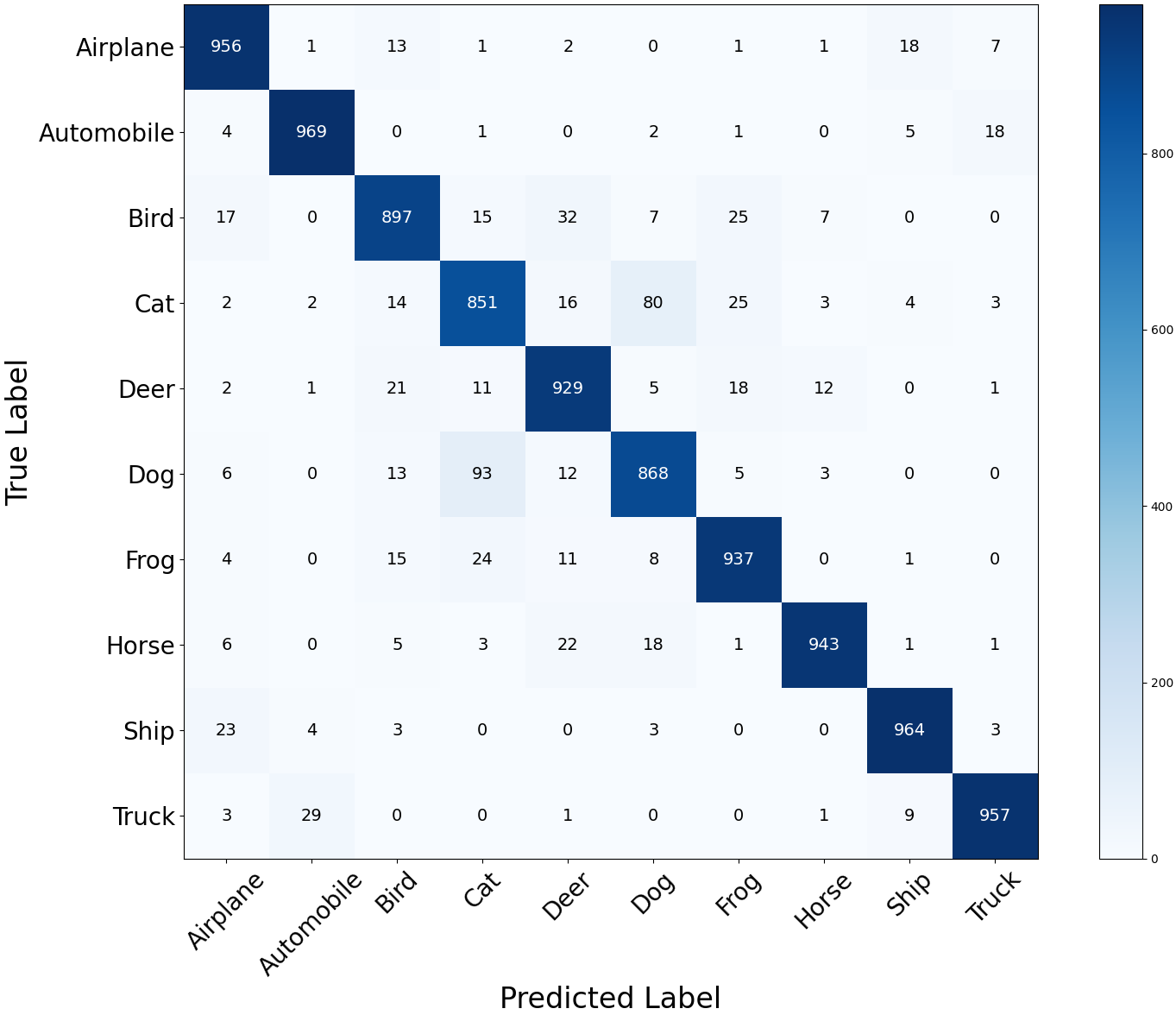}
    \vspace{-0.8em}
    \caption{Confusion matrix of prediction results on CIFAR-10 test set. This matrix illustrates that \textsc{CapRecover} can accurately reconstruct the image classes.}
    \label{fig:exp_img_classification_confusion_matrix}
    \vspace{-0.8em}
\end{figure}

Given the large number of categories in the TinyImageNet dataset, which makes detailed per-class analysis less tractable, we focus our in-depth experimental analysis on the CIFAR-10 dataset. Based on the experimental results presented in Figure \ref{fig:exp_img_classification_confusion_matrix} and Table \ref{tab:exp_img_classification}, we observe that \textsc{CapRecover} demonstrates strong performance in reconstructing the classification labels of objects from intermediate image features on the CIFAR10 test set. The confusion matrix (Figure \ref{fig:exp_img_classification_confusion_matrix}) illustrates clear and distinct diagonal patterns, indicating that predictions generally align closely with true labels. Most object categories such as ``Automobile'', ``Ship'', and ``Truck'' achieve very high correct prediction counts, approaching nearly perfect reconstruction accuracy.

Further quantitative analysis in Table \ref{tab:exp_img_classification} confirms these observations. \textsc{CapRecover} achieves consistently high precision, recall, and F1-scores across all ten classes, with scores predominantly above 0.90. The ``Truck'' and ``Ship'' categories achieve particularly high scores (F1-scores of 0.96), demonstrating especially robust performance. Even categories with slightly lower performance, such as ``Cat'' (F1-score of 0.85), remain sufficiently accurate to confirm the model's effectiveness.

Overall, these experimental results indicate that \textsc{CapRecover} effectively reconstructs object classifications from intermediate features, highlighting its potential to successfully exploit feature leakage vulnerabilities in image recognition scenarios.

\section{Discussion on Potential Defense Mechanisms}
\label{sec:protection_approach}

\subsection{Noise-Based Feature Obfuscation}

\begin{table}[t]
    \centering
    \caption{Evaluation results of \textsc{CapRecover} attacking ResNet50 with/without additional noise.}
    \label{tab:exp_add_noise}
    \vspace{-1.0em}
    \begin{threeparttable}
        \sizefive
        \setlength{\tabcolsep}{7pt}
        \renewcommand{\arraystretch}{1.2}
        \begin{tabular}{cccccc}
            \toprule
            Dataset & Middle layer & layer1 & layer2 & layer3 & layer4 \\
            \midrule
            \multirow{2}{*}{COCO2017} & w/o noise & 0.24 & 0.51       & 0.58        & 0.62         \\
            & w/ noise & 0.49 & 0.03       & 0.02        & 0.05    \\
            \bottomrule
        \end{tabular}
    \end{threeparttable}
    \vspace{-1.0em}
\end{table}

To protect intermediate representations in split DNN deployments, we propose a lightweight noise-based defense mechanism that introduces random Gaussian noise into the intermediate features during inference. While this technique effectively degrades feature inversion attacks, its practical feasibility hinges on low communication and computation overhead, especially in edge–cloud settings.

\textbf{Local-only noise handling.}  
To ensure deployment efficiency, our design ensures that both the injection and noise removal are performed entirely on the client-side (i.e., the edge device). Specifically, for any intermediate feature $F^{(i)}$, the edge device generates a random noise vector $\epsilon^{(i)} \sim \mathcal{N}(0, \sigma^2)$, computes the obfuscated representation $\tilde{F}^{(i)} = F^{(i)} + \epsilon^{(i)}$, and then removes the noise in the subsequent layer before transmitting the result to the cloud:
\[
F^{(i+1)} = g(\tilde{F}^{(i)} - \epsilon^{(i)}) = g(F^{(i)}).
\]
The noise is neither stored nor transmitted—thus, incurring \textit{no additional communication cost} and avoiding the need for synchronization with the cloud.

\textbf{Negligible computational overhead.}  
The only extra computation required is sampling from a standard Gaussian distribution and applying addition/subtraction operations—both of which are lightweight and can be efficiently executed on modern edge hardware (e.g., CPUs or NPUs). In our measurements, the time overhead introduced per inference was negligible ($<1\%$ relative increase), making this defense practical for real-time applications.

\textbf{Security benefit.}  
By ensuring that the intermediate features transmitted to the server are never raw (i.e., always processed), attackers who intercept these representations cannot reconstruct accurate semantic content without knowledge of the locally generated $\epsilon^{(i)}$. Moreover, since the noise is regenerated for each image, even partial leaks from one instance do not compromise others.

Overall, his defense achieves a strong trade-off between privacy protection and deployment practicality. It requires no retraining, with no changes to final predictions. It can be readily integrated into edge-side inference pipelines with minimal modification.

\subsection{Potential for Homomorphic Encryption}

Homomorphic Encryption (HE) \cite{wiki_def_he} represents a promising cryptographic approach to mitigating privacy risks associated with Vision-Language Models (VLMs). In typical VLM deployments, intermediate image features generated by the visual encoder are frequently transmitted between client devices and remote servers, creating opportunities for attackers to intercept and exploit these representations to reconstruct sensitive textual information, such as image captions. By encrypting these intermediate features homomorphically, HE enables operations to be conducted directly on encrypted data without revealing the underlying plaintext features, thereby significantly reducing the risk of privacy leakage.

The primary advantage of employing HE in VLM scenarios lies in its capability to ensure data confidentiality even when intermediate features are intercepted during transmission or while temporarily stored. Attackers accessing encrypted features would find it computationally infeasible to derive meaningful information without the appropriate decryption keys, effectively safeguarding sensitive textual descriptions embedded within the features.

There are practical precedents demonstrating the feasibility of HE in protecting model parameters and gradients within federated learning scenarios. For example, NVIDIA researchers successfully integrated homomorphic encryption with XGBoost \cite{xu2025secure}, employing CUDA acceleration to achieve efficient privacy-preserving federated learning. Such cases provide encouraging evidence that similar strategies could protect privacy from intermediate features, thus securing textual information from feature inversion attacks.

\section{Conclusion}
\label{sec:conclusion}

%

\par\noindent In this paper, we focus on the cross-modality feature inversion attack, proposing \textsc{CapRecover}, a generic framework that reconstructs image captions and classification labels directly from leaked intermediate image features. By leveraging a feature projection module and a feature-to-text alignment mechanism, \textsc{CapRecover} effectively recovers semantic information—even when using features from the final linear projection layer of the visual encoder. Our extensive experiments demonstrate \textsc{CapRecover}'s effectiveness across multiple datasets and models. Furthermore, we propose an effective protection approach without additional training costs, thereby efficiently preventing attackers from reconstructing sensitive image information. 

    

\bibliographystyle{ACM-Reference-Format}
\balance
\bibliography{ref}


\begin{thebibliography}{32}


\ifx \showCODEN    \undefined \def \showCODEN     #1{\unskip}     \fi
\ifx \showDOI      \undefined \def \showDOI       #1{#1}\fi
\ifx \showISBNx    \undefined \def \showISBNx     #1{\unskip}     \fi
\ifx \showISBNxiii \undefined \def \showISBNxiii  #1{\unskip}     \fi
\ifx \showISSN     \undefined \def \showISSN      #1{\unskip}     \fi
\ifx \showLCCN     \undefined \def \showLCCN      #1{\unskip}     \fi
\ifx \shownote     \undefined \def \shownote      #1{#1}          \fi
\ifx \showarticletitle \undefined \def \showarticletitle #1{#1}   \fi
\ifx \showURL      \undefined \def \showURL       {\relax}        \fi
\providecommand\bibfield[2]{#2}
\providecommand\bibinfo[2]{#2}
\providecommand\natexlab[1]{#1}
\providecommand\showeprint[2][]{arXiv:#2}

\bibitem[AI({[n.\,d.]})]%
        {stable_diffusion}
\bibfield{author}{\bibinfo{person}{Stability AI}.} \bibinfo{year}{[n.\,d.]}\natexlab{}.
\newblock \bibinfo{booktitle}{\emph{Activating humanity's potential through generative AI}}.
\newblock
\newblock
\shownote{\url{https://stability.ai/}}.


\bibitem[Deng et~al\mbox{.}(2009)]%
        {dst_imagenet}
\bibfield{author}{\bibinfo{person}{Jia Deng}, \bibinfo{person}{Wei Dong}, \bibinfo{person}{Richard Socher}, \bibinfo{person}{Li-Jia Li}, \bibinfo{person}{Kai Li}, {and} \bibinfo{person}{Li Fei-Fei}.} \bibinfo{year}{2009}\natexlab{}.
\newblock \showarticletitle{ImageNet: A large-scale hierarchical image database}. In \bibinfo{booktitle}{\emph{2009 IEEE Conference on Computer Vision and Pattern Recognition}}. \bibinfo{pages}{248--255}.
\newblock
\urldef\tempurl%
\url{https://doi.org/10.1109/CVPR.2009.5206848}
\showDOI{\tempurl}


\bibitem[Dosovitskiy(2020)]%
        {dosovitskiy2020image}
\bibfield{author}{\bibinfo{person}{Alexey Dosovitskiy}.} \bibinfo{year}{2020}\natexlab{}.
\newblock \showarticletitle{An image is worth 16x16 words: Transformers for image recognition at scale}.
\newblock \bibinfo{journal}{\emph{arXiv preprint arXiv:2010.11929}} (\bibinfo{year}{2020}).
\newblock


\bibitem[Gong et~al\mbox{.}(2023)]%
        {gong2023figstep}
\bibfield{author}{\bibinfo{person}{Yichen Gong}, \bibinfo{person}{Delong Ran}, \bibinfo{person}{Jinyuan Liu}, \bibinfo{person}{Conglei Wang}, \bibinfo{person}{Tianshuo Cong}, \bibinfo{person}{Anyu Wang}, \bibinfo{person}{Sisi Duan}, {and} \bibinfo{person}{Xiaoyun Wang}.} \bibinfo{year}{2023}\natexlab{}.
\newblock \bibinfo{title}{FigStep: Jailbreaking Large Vision-language Models via Typographic Visual Prompts}.
\newblock
\newblock
\showeprint[arxiv]{2311.05608}~[cs.CR]


\bibitem[He et~al\mbox{.}(2016)]%
        {he2016deep}
\bibfield{author}{\bibinfo{person}{Kaiming He}, \bibinfo{person}{Xiangyu Zhang}, \bibinfo{person}{Shaoqing Ren}, {and} \bibinfo{person}{Jian Sun}.} \bibinfo{year}{2016}\natexlab{}.
\newblock \showarticletitle{Deep residual learning for image recognition}. In \bibinfo{booktitle}{\emph{Proceedings of the IEEE conference on computer vision and pattern recognition}}. \bibinfo{pages}{770--778}.
\newblock


\bibitem[He et~al\mbox{.}(2024)]%
        {he2024large}
\bibfield{author}{\bibinfo{person}{Ying He}, \bibinfo{person}{Jingcheng Fang}, \bibinfo{person}{F~Richard Yu}, {and} \bibinfo{person}{Victor~C Leung}.} \bibinfo{year}{2024}\natexlab{}.
\newblock \showarticletitle{Large language models (LLMs) inference offloading and resource allocation in cloud-edge computing: An active inference approach}.
\newblock \bibinfo{journal}{\emph{IEEE Transactions on Mobile Computing}} (\bibinfo{year}{2024}).
\newblock


\bibitem[He et~al\mbox{.}(2019)]%
        {he2019model}
\bibfield{author}{\bibinfo{person}{Zecheng He}, \bibinfo{person}{Tianwei Zhang}, {and} \bibinfo{person}{Ruby~B Lee}.} \bibinfo{year}{2019}\natexlab{}.
\newblock \showarticletitle{Model inversion attacks against collaborative inference}. In \bibinfo{booktitle}{\emph{Proceedings of the 35th Annual Computer Security Applications Conference (ACSAC)}}. \bibinfo{pages}{148--162}.
\newblock


\bibitem[Hodosh et~al\mbox{.}(2013)]%
        {dst_flickr8k}
\bibfield{author}{\bibinfo{person}{Micah Hodosh}, \bibinfo{person}{Peter Young}, {and} \bibinfo{person}{Julia Hockenmaier}.} \bibinfo{year}{2013}\natexlab{}.
\newblock \showarticletitle{Framing image description as a ranking task: Data, models and evaluation metrics}.
\newblock \bibinfo{journal}{\emph{Journal of Artificial Intelligence Research}} (\bibinfo{date}{Aug.} \bibinfo{year}{2013}), \bibinfo{pages}{853--899}.
\newblock
\urldef\tempurl%
\url{https://doi.org/10.1613/jair.3994}
\showDOI{\tempurl}


\bibitem[Howard et~al\mbox{.}(2019)]%
        {howard2019searching}
\bibfield{author}{\bibinfo{person}{Andrew Howard}, \bibinfo{person}{Mark Sandler}, \bibinfo{person}{Grace Chu}, \bibinfo{person}{Liang-Chieh Chen}, \bibinfo{person}{Bo Chen}, \bibinfo{person}{Mingxing Tan}, \bibinfo{person}{Weijun Wang}, \bibinfo{person}{Yukun Zhu}, \bibinfo{person}{Ruoming Pang}, \bibinfo{person}{Vijay Vasudevan}, {et~al\mbox{.}}} \bibinfo{year}{2019}\natexlab{}.
\newblock \showarticletitle{Searching for mobilenetv3}. In \bibinfo{booktitle}{\emph{Proceedings of the IEEE/CVF international conference on computer vision}}. \bibinfo{pages}{1314--1324}.
\newblock


\bibitem[Huang et~al\mbox{.}(2022)]%
        {huang2022unsupervised}
\bibfield{author}{\bibinfo{person}{Tony Huang}, \bibinfo{person}{Jack Chu}, {and} \bibinfo{person}{Fangyun Wei}.} \bibinfo{year}{2022}\natexlab{}.
\newblock \showarticletitle{Unsupervised prompt learning for vision-language models}.
\newblock \bibinfo{journal}{\emph{arXiv preprint arXiv:2204.03649}} (\bibinfo{year}{2022}).
\newblock


\bibitem[Lei et~al\mbox{.}(2023)]%
        {lei2023image}
\bibfield{author}{\bibinfo{person}{Shiye Lei}, \bibinfo{person}{Hao Chen}, \bibinfo{person}{Sen Zhang}, \bibinfo{person}{Bo Zhao}, {and} \bibinfo{person}{Dacheng Tao}.} \bibinfo{year}{2023}\natexlab{}.
\newblock \showarticletitle{Image captions are natural prompts for text-to-image models}.
\newblock \bibinfo{journal}{\emph{arXiv preprint arXiv:2307.08526}} (\bibinfo{year}{2023}).
\newblock


\bibitem[Li et~al\mbox{.}(2023b)]%
        {blip2}
\bibfield{author}{\bibinfo{person}{Junnan Li}, \bibinfo{person}{Dongxu Li}, \bibinfo{person}{Silvio Savarese}, {and} \bibinfo{person}{Steven Hoi}.} \bibinfo{year}{2023}\natexlab{b}.
\newblock \showarticletitle{Blip-2: Bootstrapping language-image pre-training with frozen image encoders and large language models}. In \bibinfo{booktitle}{\emph{International conference on machine learning}}. PMLR, \bibinfo{pages}{19730--19742}.
\newblock


\bibitem[Li et~al\mbox{.}(2023a)]%
        {li2023rethinking}
\bibfield{author}{\bibinfo{person}{Yanyu Li}, \bibinfo{person}{Ju Hu}, \bibinfo{person}{Yang Wen}, \bibinfo{person}{Georgios Evangelidis}, \bibinfo{person}{Kamyar Salahi}, \bibinfo{person}{Yanzhi Wang}, \bibinfo{person}{Sergey Tulyakov}, {and} \bibinfo{person}{Jian Ren}.} \bibinfo{year}{2023}\natexlab{a}.
\newblock \showarticletitle{Rethinking vision transformers for mobilenet size and speed}. In \bibinfo{booktitle}{\emph{Proceedings of the IEEE/CVF international conference on computer vision}}. \bibinfo{pages}{16889--16900}.
\newblock


\bibitem[Li et~al\mbox{.}(2022)]%
        {li2022efficientformer}
\bibfield{author}{\bibinfo{person}{Yanyu Li}, \bibinfo{person}{Geng Yuan}, \bibinfo{person}{Yang Wen}, \bibinfo{person}{Ju Hu}, \bibinfo{person}{Georgios Evangelidis}, \bibinfo{person}{Sergey Tulyakov}, \bibinfo{person}{Yanzhi Wang}, {and} \bibinfo{person}{Jian Ren}.} \bibinfo{year}{2022}\natexlab{}.
\newblock \showarticletitle{Efficientformer: Vision transformers at mobilenet speed}.
\newblock \bibinfo{journal}{\emph{Advances in Neural Information Processing Systems}}  \bibinfo{volume}{35} (\bibinfo{year}{2022}), \bibinfo{pages}{12934--12949}.
\newblock


\bibitem[Lin et~al\mbox{.}(2014)]%
        {lin2014microsoft}
\bibfield{author}{\bibinfo{person}{Tsung-Yi Lin}, \bibinfo{person}{Michael Maire}, \bibinfo{person}{Serge Belongie}, \bibinfo{person}{James Hays}, \bibinfo{person}{Pietro Perona}, \bibinfo{person}{Deva Ramanan}, \bibinfo{person}{Piotr Doll{\'a}r}, {and} \bibinfo{person}{C~Lawrence Zitnick}.} \bibinfo{year}{2014}\natexlab{}.
\newblock \showarticletitle{Microsoft coco: Common objects in context}. In \bibinfo{booktitle}{\emph{Computer Vision--ECCV 2014: 13th European Conference, Zurich, Switzerland, September 6-12, 2014, Proceedings, Part V 13}}. Springer, \bibinfo{pages}{740--755}.
\newblock


\bibitem[Liu et~al\mbox{.}(2023)]%
        {liu2023llava}
\bibfield{author}{\bibinfo{person}{Haotian Liu}, \bibinfo{person}{Chunyuan Li}, \bibinfo{person}{Qingyang Wu}, {and} \bibinfo{person}{Yong~Jae Lee}.} \bibinfo{year}{2023}\natexlab{}.
\newblock \bibinfo{title}{Visual Instruction Tuning}.
\newblock
\newblock


\bibitem[Lu et~al\mbox{.}(2024)]%
        {lu2024merge}
\bibfield{author}{\bibinfo{person}{Jinliang Lu}, \bibinfo{person}{Ziliang Pang}, \bibinfo{person}{Min Xiao}, \bibinfo{person}{Yaochen Zhu}, \bibinfo{person}{Rui Xia}, {and} \bibinfo{person}{Jiajun Zhang}.} \bibinfo{year}{2024}\natexlab{}.
\newblock \showarticletitle{Merge, ensemble, and cooperate! a survey on collaborative strategies in the era of large language models}.
\newblock \bibinfo{journal}{\emph{arXiv preprint arXiv:2407.06089}} (\bibinfo{year}{2024}).
\newblock


\bibitem[Luo et~al\mbox{.}(2024)]%
        {luo2024jailbreakv28k}
\bibfield{author}{\bibinfo{person}{Weidi Luo}, \bibinfo{person}{Siyuan Ma}, \bibinfo{person}{Xiaogeng Liu}, \bibinfo{person}{Xiaoyu Guo}, {and} \bibinfo{person}{Chaowei Xiao}.} \bibinfo{year}{2024}\natexlab{}.
\newblock \bibinfo{title}{JailBreakV-28K: A Benchmark for Assessing the Robustness of MultiModal Large Language Models against Jailbreak Attacks}.
\newblock
\newblock
\showeprint[arxiv]{2404.03027}~[cs.CR]


\bibitem[Mudvari et~al\mbox{.}(2024)]%
        {mudvari2024splitllm}
\bibfield{author}{\bibinfo{person}{Akrit Mudvari}, \bibinfo{person}{Yuang Jiang}, {and} \bibinfo{person}{Leandros Tassiulas}.} \bibinfo{year}{2024}\natexlab{}.
\newblock \showarticletitle{Splitllm: Collaborative inference of llms for model placement and throughput optimization}.
\newblock \bibinfo{journal}{\emph{arXiv preprint arXiv:2410.10759}} (\bibinfo{year}{2024}).
\newblock


\bibitem[Nguyen et~al\mbox{.}(2023)]%
        {nguyen2023improving}
\bibfield{author}{\bibinfo{person}{Thao Nguyen}, \bibinfo{person}{Samir~Yitzhak Gadre}, \bibinfo{person}{Gabriel Ilharco}, \bibinfo{person}{Sewoong Oh}, {and} \bibinfo{person}{Ludwig Schmidt}.} \bibinfo{year}{2023}\natexlab{}.
\newblock \showarticletitle{Improving multimodal datasets with image captioning}.
\newblock \bibinfo{journal}{\emph{Advances in neural information processing systems}}  \bibinfo{volume}{36} (\bibinfo{year}{2023}), \bibinfo{pages}{22047--22069}.
\newblock


\bibitem[OpenAI(2024)]%
        {openai_gpt4o}
\bibfield{author}{\bibinfo{person}{OpenAI}.} \bibinfo{year}{2024}\natexlab{}.
\newblock \bibinfo{booktitle}{\emph{Hello GPT-4o}}.
\newblock
\newblock
\shownote{\url{https://openai.com/index/hello-gpt-4o/}}.


\bibitem[Radford et~al\mbox{.}(2021)]%
        {clip}
\bibfield{author}{\bibinfo{person}{Alec Radford}, \bibinfo{person}{Jong~Wook Kim}, \bibinfo{person}{Chris Hallacy}, \bibinfo{person}{Aditya Ramesh}, \bibinfo{person}{Gabriel Goh}, \bibinfo{person}{Sandhini Agarwal}, \bibinfo{person}{Girish Sastry}, \bibinfo{person}{Amanda Askell}, \bibinfo{person}{Pamela Mishkin}, \bibinfo{person}{Jack Clark}, {et~al\mbox{.}}} \bibinfo{year}{2021}\natexlab{}.
\newblock \showarticletitle{Learning transferable visual models from natural language supervision}. In \bibinfo{booktitle}{\emph{International conference on machine learning}}. PMLR, \bibinfo{pages}{8748--8763}.
\newblock


\bibitem[Sandler et~al\mbox{.}(2018)]%
        {sandler2018mobilenetv2}
\bibfield{author}{\bibinfo{person}{Mark Sandler}, \bibinfo{person}{Andrew Howard}, \bibinfo{person}{Menglong Zhu}, \bibinfo{person}{Andrey Zhmoginov}, {and} \bibinfo{person}{Liang-Chieh Chen}.} \bibinfo{year}{2018}\natexlab{}.
\newblock \showarticletitle{Mobilenetv2: Inverted residuals and linear bottlenecks}. In \bibinfo{booktitle}{\emph{Proceedings of the IEEE conference on computer vision and pattern recognition}}. \bibinfo{pages}{4510--4520}.
\newblock


\bibitem[Shayegani et~al\mbox{.}(2023)]%
        {shayegani2023plug}
\bibfield{author}{\bibinfo{person}{Erfan Shayegani}, \bibinfo{person}{Yue Dong}, {and} \bibinfo{person}{Nael Abu-Ghazaleh}.} \bibinfo{year}{2023}\natexlab{}.
\newblock \showarticletitle{Plug and Pray: Exploiting off-the-shelf components of Multi-Modal Models}.
\newblock \bibinfo{journal}{\emph{arXiv preprint arXiv:2307.14539}} (\bibinfo{year}{2023}).
\newblock


\bibitem[Shen et~al\mbox{.}(2024)]%
        {2024_usenix_prompt_stealing_attack}
\bibfield{author}{\bibinfo{person}{Xinyue Shen}, \bibinfo{person}{Yiting Qu}, \bibinfo{person}{Michael Backes}, {and} \bibinfo{person}{Yang Zhang}.} \bibinfo{year}{2024}\natexlab{}.
\newblock \showarticletitle{{Prompt Stealing Attacks Against Text-to-Image Generation Models}}. In \bibinfo{booktitle}{\emph{{USENIX Security Symposium (USENIX Security)}}}. \bibinfo{publisher}{USENIX}.
\newblock


\bibitem[Vasu et~al\mbox{.}(2023)]%
        {vasu2023fastvit}
\bibfield{author}{\bibinfo{person}{Pavan Kumar~Anasosalu Vasu}, \bibinfo{person}{James Gabriel}, \bibinfo{person}{Jeff Zhu}, \bibinfo{person}{Oncel Tuzel}, {and} \bibinfo{person}{Anurag Ranjan}.} \bibinfo{year}{2023}\natexlab{}.
\newblock \showarticletitle{Fastvit: A fast hybrid vision transformer using structural reparameterization}. In \bibinfo{booktitle}{\emph{Proceedings of the IEEE/CVF international conference on computer vision}}. \bibinfo{pages}{5785--5795}.
\newblock


\bibitem[Wikipedia({[n.\,d.]})]%
        {wiki_def_he}
\bibfield{author}{\bibinfo{person}{Wikipedia}.} \bibinfo{year}{[n.\,d.]}\natexlab{}.
\newblock \bibinfo{booktitle}{\emph{Homomorphic encryption}}.
\newblock
\newblock
\shownote{\url{https://en.wikipedia.org/wiki/Homomorphic_encryption}}.


\bibitem[Xu et~al\mbox{.}(2024)]%
        {xu2024stealthy}
\bibfield{author}{\bibinfo{person}{Xiaoyang Xu}, \bibinfo{person}{Mengda Yang}, \bibinfo{person}{Wenzhe Yi}, \bibinfo{person}{Ziang Li}, \bibinfo{person}{Juan Wang}, \bibinfo{person}{Hongxin Hu}, \bibinfo{person}{Yong Zhuang}, {and} \bibinfo{person}{Yaxin Liu}.} \bibinfo{year}{2024}\natexlab{}.
\newblock \showarticletitle{A Stealthy Wrongdoer: Feature-Oriented Reconstruction Attack against Split Learning}. In \bibinfo{booktitle}{\emph{Proceedings of the IEEE/CVF Conference on Computer Vision and Pattern Recognition}}. \bibinfo{pages}{12130--12139}.
\newblock


\bibitem[Xu et~al\mbox{.}(2025)]%
        {xu2025secure}
\bibfield{author}{\bibinfo{person}{Ziyue Xu}, \bibinfo{person}{Yuan-Ting Hsieh}, \bibinfo{person}{Zhihong Zhang}, \bibinfo{person}{Holger~R Roth}, \bibinfo{person}{Chester Chen}, \bibinfo{person}{Yan Cheng}, {and} \bibinfo{person}{Andrew Feng}.} \bibinfo{year}{2025}\natexlab{}.
\newblock \showarticletitle{Secure Federated XGBoost with CUDA-accelerated Homomorphic Encryption via NVIDIA FLARE}.
\newblock \bibinfo{journal}{\emph{arXiv preprint arXiv:2504.03909}} (\bibinfo{year}{2025}).
\newblock


\bibitem[Yang et~al\mbox{.}(2024)]%
        {yang2024qwen2}
\bibfield{author}{\bibinfo{person}{An Yang}, \bibinfo{person}{Baosong Yang}, \bibinfo{person}{Beichen Zhang}, \bibinfo{person}{Binyuan Hui}, \bibinfo{person}{Bo Zheng}, \bibinfo{person}{Bowen Yu}, \bibinfo{person}{Chengyuan Li}, \bibinfo{person}{Dayiheng Liu}, \bibinfo{person}{Fei Huang}, \bibinfo{person}{Haoran Wei}, {et~al\mbox{.}}} \bibinfo{year}{2024}\natexlab{}.
\newblock \showarticletitle{Qwen2. 5 technical report}.
\newblock \bibinfo{journal}{\emph{arXiv preprint arXiv:2412.15115}} (\bibinfo{year}{2024}).
\newblock


\bibitem[Zhang et~al\mbox{.}(2024)]%
        {zhang2024edgeshard}
\bibfield{author}{\bibinfo{person}{Mingjin Zhang}, \bibinfo{person}{Xiaoming Shen}, \bibinfo{person}{Jiannong Cao}, \bibinfo{person}{Zeyang Cui}, {and} \bibinfo{person}{Shan Jiang}.} \bibinfo{year}{2024}\natexlab{}.
\newblock \showarticletitle{Edgeshard: Efficient llm inference via collaborative edge computing}.
\newblock \bibinfo{journal}{\emph{IEEE Internet of Things Journal}} (\bibinfo{year}{2024}).
\newblock


\bibitem[Zhu et~al\mbox{.}(2025)]%
        {zhu2025passiveinfer}
\bibfield{author}{\bibinfo{person}{Xiaochen Zhu}, \bibinfo{person}{Xinjian Luo}, \bibinfo{person}{Yuncheng Wu}, \bibinfo{person}{Yangfan Jiang}, \bibinfo{person}{Xiaokui Xiao}, {and} \bibinfo{person}{Beng~Chin Ooi}.} \bibinfo{year}{2025}\natexlab{}.
\newblock \showarticletitle{Passive Inference Attacks on Split Learning via Adversarial Regularization}. In \bibinfo{booktitle}{\emph{Network and Distributed System Security (NDSS) Symposium 2025}}.
\newblock


\end{thebibliography}

\end{document}